\documentclass[10pt,twocolumn,letterpaper]{article}

\usepackage{iccv}              

\definecolor{iccvblue}{rgb}{0.21,0.49,0.74}
\usepackage[pagebackref,breaklinks,colorlinks,allcolors=iccvblue]{hyperref}

\usepackage{times}
\usepackage{graphicx}
\usepackage{amssymb}
\usepackage{epstopdf}
\usepackage{multirow}
\usepackage{setspace}  
\usepackage{verbatim}
\usepackage{amsmath}
\usepackage[normalem]{ulem}
\usepackage[margin=1in]{geometry}  
\usepackage{sectsty}
\usepackage{rotating}
\usepackage{paralist}
\usepackage{enumitem}

\usepackage{titlesec}
\usepackage{bm}
\usepackage{bbm}
\usepackage{sidecap}
\usepackage{booktabs}

\usepackage[ruled]{algorithm2e}
\usepackage[accsupp]{axessibility}

\def\paperID{5488} 
\def\confName{ICCV}
\def\confYear{2025}

\newcommand{\CUT}[1]{}
\newcommand{\jimmy}{}

\def\real{\mathbb{R}}

\newcommand{\bdelta}{\bm{\delta}}
\newcommand{\bDelta}{\bm{\Delta}}

\newcommand{\bb}{\mathbf{b}}

\newcommand{\bx}{\mathbf{x}}

\newcommand{\bz}{\mathbf{z}}

\newcommand{\bI}{\mathbf{I}}

\title{Temporal Unlearnable Examples: Preventing Personal Video Data from Unauthorized Exploitation by Object Tracking}

\author{
  Qiangqiang Wu\textsuperscript{1}\thanks{Equal contribution. \textsuperscript{$\dagger$}Corresponding author.} \quad
  Yi Yu\textsuperscript{2}\footnotemark[1] \quad
  Chenqi Kong\textsuperscript{2$\dagger$} \quad
  Ziquan Liu\textsuperscript{3} \quad
  Jia Wan\textsuperscript{4} \\\quad
  Haoliang Li\textsuperscript{1} \quad
  Alex C. Kot\textsuperscript{2} \quad
  Antoni B. Chan\textsuperscript{1} \\
  \textsuperscript{1}City University of Hong Kong \quad
  \textsuperscript{2}ROSE Lab, Nanyang Technological University \quad \\
  \textsuperscript{3}Queen Mary University of London \quad
  \textsuperscript{4}Harbin Institute of Technology, Shenzhen\\
  {\tt\small qiangqwu2-c@my.cityu.edu.hk, \{yuyi0010, chenqi.kong, eackot\}@ntu.edu.sg,}\\ {\tt\small ziquan.liu@qmul.ac.uk, jiawan1998@gmail.com, \{haoliang.li, abchan\}@cityu.edu.hk}
}

\begin{document}
\maketitle
\begin{abstract}
With the rise of social media, vast amounts of user-uploaded videos (e.g., YouTube) are utilized as training data for Visual Object Tracking (VOT). However, the VOT community has largely overlooked video data-privacy issues, as many private videos have been collected and used for training commercial models without authorization. To alleviate these issues, this paper presents the first investigation on preventing personal video data from unauthorized exploitation by deep trackers. Existing methods for preventing unauthorized data use primarily focus on image-based tasks (e.g., image classification), directly applying them to videos reveals several limitations, including inefficiency, limited effectiveness, and poor generalizability.  
To address these issues, we propose a novel generative framework for generating Temporal Unlearnable Examples (TUEs), and whose efficient computation makes it scalable for usage on large-scale video datasets. The trackers trained w/ TUEs heavily rely on unlearnable noises for temporal matching, ignoring the original data structure and thus ensuring training video data-privacy. To enhance the effectiveness of TUEs, we introduce a temporal contrastive loss, which further corrupts the learning of existing trackers when using our TUEs for training. Extensive experiments demonstrate that our approach achieves state-of-the-art performance in video data-privacy protection, with strong transferability across 
VOT models, datasets, and temporal matching tasks.
\end{abstract}    
\section{Introduction}
\label{sec:intro}

\begin{figure}[t]
  \centering
  \includegraphics[width=0.9\linewidth]{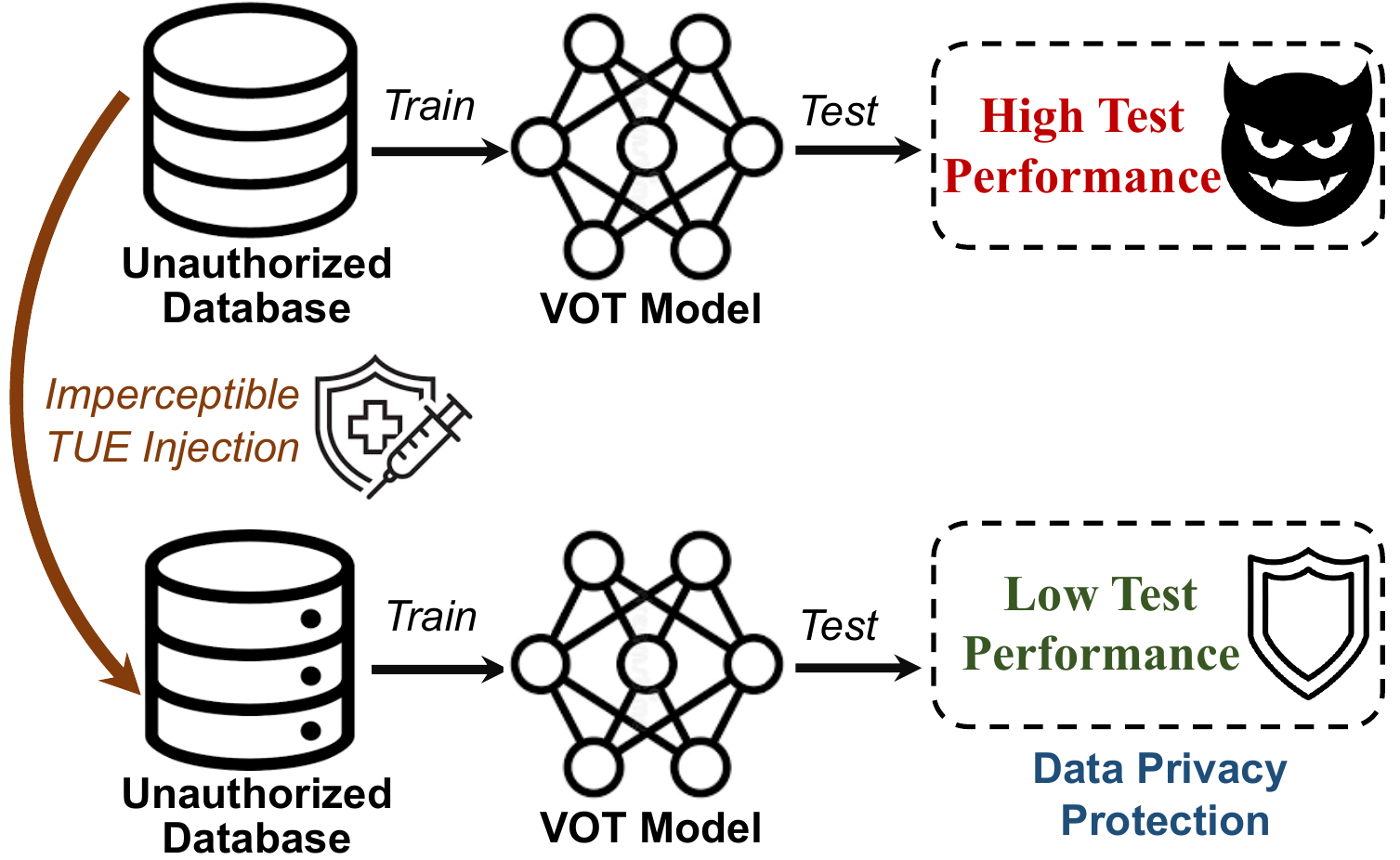}
   \caption{Illustration of our TUEs for preventing video data from unauthorized exploitation by deep VOT models. Adding imperceptible TUEs to training videos limits deep VOT models to only learning shortcuts information, resulting in poor generalization and degraded test performance. 
    }
   \label{teaser}
\end{figure}

{Visual Object Tracking (VOT) estimates target bounding boxes in each video frame based on the initial target state, playing a key role in applications like intelligent surveillance and autonomous driving. Recent advances in deep trackers, driven by large-scale training videos from the internet, have significantly improved VOT performance. However, concerns over unauthorized data exploitation by Deep Neural Networks (DNNs) are growing. For instance, personal videos uploaded to social media (\textit{e.g.,} YouTube) may be used for VOT training without consent, raising privacy and copyright issues. Large-scale VOT datasets like TrackingNet \cite{trackingnet}, LaSOT \cite{lasot,lasotext}, and GOT-10k \cite{got10k} primarily consist of such user-uploaded videos. Protecting sensitive trajectories—such as those of individuals, vehicles, and military assets—requires preventing unauthorized use of tracking data. Hence, safeguarding video data from exploitation in VOT training is essential.}

\jimmy{In the context of 2D images, Unlearnable Examples (UEs) \cite{em,rem,yu2024purify,yu2025mtlue} is a typical solution to protect private data from unauthorized exploitation by DNNs. UEs methods add imperceptible perturbations (\textit{i.e.,} bounded noises) to the training images to hinder models from extracting useful information from the poisoned training data, thereby resulting in poor testing performance. 
{Extending image-task UEs to videos remains underexplored. While uniform perturbations across frames may effectively adapt image-task UEs for video classification tasks like action recognition, they struggle with temporal matching tasks such as VOT and Video Object Segmentation (VOS). Our paper focuses on learning UEs for temporal matching, with VOT as a foundational task. Protecting VOT is crucial for safeguarding sensitive trajectories (e.g., missiles, vehicles, and individuals). Optimizing UEs for the basic VOT task could also improve their transferability to advanced temporal matching tasks such as VOS and long-term point tracking.}
}

{Extending image-task UEs to temporal matching tasks presents several challenges: (1) Video data have higher resolution and more frames, making UEs generation computationally intensive; (2) Unlike image classification, VOT relies on temporal matching across frames, with target objects changing in scale, complicating the design of UEs that support scale-invariant matching; (3) Existing UEs struggle to transfer across different tracking models, datasets, and matching tasks, limiting their effectiveness for data privacy in VOT. 
}

{To address the above challenges, we introduce Temporal Unlearnable Examples (TUEs) for video data (see Fig.~\ref{teaser}). By injecting imperceptible TUEs, deep VOT models learn only limited information from the training set, resulting in poor generalization and degraded test performance.
We propose a novel generative framework in which a generator is trained to produce TUEs that disrupt temporal matching in VOT models. These perturbations remain imperceptible to the human eye and do not compromise the data utility for human consumption. Compared to the traditional iterative EM approach \cite{em} used for image UEs, our framework significantly improves efficiency—achieving 4× faster training speed and 28× greater parameter efficiency on the GOT-10k \cite{got10k} dataset—making it highly scalable for large video datasets.
Additionally, we introduce a temporal contrastive loss (TCL) to encourage trackers to rely more on the generated TUEs for temporal matching, thereby further enhancing the privacy protection.}

{We assess the transferability of our TUEs through experiments across various VOT models and datasets. TUEs trained on the simple SiamFC \cite{SiamFC} tracker can effectively degrade the performance of state-of-the-art deep trackers with complex architectures, such as ViT \cite{ViT,zhang2022hivit} and ResNet \cite{resnet}. Moreover, a generator trained on a source dataset like GOT-10k \cite{got10k} can be used for zero-shot TUE generation on unseen video datasets, without retraining. Finally, we show that our TUEs are task-transferable and perform well in other temporal matching tasks, such as  video object segmentation.}

In summary, the main contributions of our work are:
$\indent\bullet$ To the best of our knowledge, we are the first to investigate preventing unauthorized video exploitation for VOT. Since none exists, we apply off-the-shelf image-task UEs to videos as baselines for VOT with specific designs, which reveal several limitations, \textit{e.g.,} inefficiency, limited effectiveness, and poor generalizability.

$\bullet$ We propose a new generative framework to generate Temporal Unlearnable Examples (TUEs), which can effectively corrupt the temporal matching learning of VOT models. Our method achieves state-of-the-art performance in video data privacy protection and shows strong transferability across various trackers, datasets, and matching tasks.

$\bullet$ We introduce a Temporal Contrastive Loss (TCL) that encourages trackers to rely more on the generated TUEs for temporal matching learning, which further degrades the tracking performance while preserving the data privacy of training data.
\section{Related Work}
\label{sec:RelatedWork}
\noindent\textbf{Visual Object Tracking (VOT)}
{predicts target bounding boxes in each frame based on the initial target state. Early correlation-filter (CF) trackers \cite{KCF,BACF,Staple,DSST,SRDCF,CSRDCF} were successful due to their performance and speed. With the rise of deep learning \cite{alex,resnet}, CF trackers \cite{dsnet,ECO,CCOT,robust_cf_journal,CFNet,DCFNet,DeepSRDCF,liu2022new,wu2019dsnet} began incorporating deep features for VOT. SiamFC \cite{SiamFC} and SINT \cite{SINT} introduced deep Siamese networks for end-to-end VOT, leading to improvements in transformer tracking \cite{roam,ostrack,MixFormer,aqatrack}, online memory design \cite{memtrack,updatenet}, architecture design \cite{siamrpn,SiamRPN_plus,transt,hong2024onetracker,gao2023generalized,wu2023scalable}, and new learning paradigms \cite{dropmae, chen2023seqtrack, hiptrack,artrack, artrack_v2, diffusiontrack,wu2021meta}. These trackers rely on large-scale datasets like GOT-10k \cite{got10k} and LaSOT \cite{lasot,lasotext}, often sourced from user-uploaded social media videos. However, privacy concerns in VOT remain largely overlooked and demand urgent attention.
{Recently, several adversarial attacks have targeted deep trackers \cite{spark,sta,chen2020one,liang2020efficient,9142255,jia2020robust,yan2020hijacking}, such as generating temporally transferable perturbations \cite{nakka2020temporally} and introducing adversarial loss to reduce heatmap hot regions \cite{yan2020cooling}. Additionally, VOT models' vulnerability to backdoor attacks has been highlighted \cite{li2022few, huang2024badtrack}. However, these methods often require access to both training and testing data or model parameters, which can be impractical. In contrast, our approach focuses on protecting training video data from unauthorized use, requiring access only to the training data, making it more feasible.}


\vspace{1mm}
\noindent\textbf{Unlearnable Examples (UEs).}
{Data privacy in 2D images has been widely studied, with traditional methods focusing on preventing models from leaking sensitive information \cite{shokri2015privacy, abadi2016deep, phan2016differential, shokri2017membership, shan2020fawkes,lin2024hidemia,DBLP:journals/patterns/LinTWLJWY24,li2023memory,li2024cascade,li2025contrast,li2025hyman,li2025surveyunlearnabledata,li2025kbs}. UEs \cite{dc, em,lin2024safeguarding,meng2024semantic,yu2024unlearnable} have recently emerged, where bounded perturbations (\textit{e.g.,} $\Vert \boldsymbol{\delta} \Vert_{\infty} \le \frac{8}{255}$) are added to training data, preserving labels while degrading model performance. This perturbative poisoning method \cite{iss} shows promise for data protection, causing models trained on such data to perform near-randomly on clean test data. Techniques like EM \cite{em}, NTGA \cite{ntga}, TAP \cite{tap}, REM \cite{rem}, LSP \cite{lsp}, and OPS \cite{ops} offer various strategies for generating protective perturbations. While UEs have been explored in tasks like image classification, segmentation, and point cloud classification \cite{sun2024unseg, wang2024unlearnable}, their application in tasks like VOT matching remains unexplored. We aim to bridge this gap by exploring effective UEs for template matching across video frames.}


\section{Methodology}
\jimmy{
We first review the preliminaries in \S\ref{sec:preli},  and introduce UE baselines in \S\ref{sec:baseline}. To overcome UE's limitations, we propose a generative framework for Temporal Unlearnable Examples (TUEs) in  \S\ref{sec:gen_framework}. We further present Temporal Contrastive Learning to enhance TUEs in   \S\ref{sec:tcl}. Finally, we summarize the overall pipeline in \S\ref{sec:application}. 
}

\subsection{Preliminaries}\label{sec:preli}

We briefly introduce traditional UEs algorithms and the temporal matching pipeline in VOT.

\noindent\textbf{Error-Minimizing (EM) Learning.} Given a clean image dataset $\mathcal{D}_{c}=\{(\bI_{i}, {y}_i)\}_{i=1}^{m}$ with $m$ clean images, EM \cite{em} aims to learn sample-wise UEs on $\mathcal{D}_{c}$ that will  degrade a classifier $F_{\theta}$ on the testing set:
\begin{equation}
\label{bilevel}
\min_{\mathbf{\theta}}{\mathbb{E}}_{(\bI, y)\sim\mathcal{D}_{c}}\big[\min_{\bdelta\in\bDelta}\mathcal{L}_{c}(F_{\theta}(\bI+\bdelta),y)\big],
\end{equation}
where $\mathcal{L}_c$ is the cross-entropy loss, $y$ is the class label of $\bI \!\!\in\!\! \real^{H\!\times\! W\!\times\! C}$, $\bdelta \!\!\in\!\! \bDelta \!\!\subset\!\! \real^{H\!\times\! W\!\times\! C}$ is the imperceptible noises, and $\bDelta$ is the feasible region. 
%
Typically, the noise $\bdelta$ is $L_{p}$-norm bounded, \textit{i.e.,} $\bDelta \!=\! \{\bdelta \ \!|\!\  ||\bdelta||_{p}\!\leq\!\sigma\}$, where $\sigma$ is small such that the noise is imperceptible. The goal is for the noise $\bdelta$ to make the original image $\bI$ unlearnable, thus $\hat{\bI} = \bI+\bdelta$ is denoted as the UEs. 
The formulation in (\ref{bilevel}) 
involves both inner and outer optimization processes. The outer optimization fixes the perturbations and updates the  parameters $\mathbf{\theta}$ of the classifier $F_{\theta}$ by minimizing the loss $\mathcal{L}_c$. In the inner optimization, the classifier is fixed while the perturbations $\bdelta$ are updated to also minimize the loss (making each training sample ``easier" for the classifier). These optimizations are performed alternately during training. In each inner optimization, EM uses Projected Gradient Descent (PGD) \cite{pgd} to iteratively update the perturbations $\bdelta$ over $T$ steps. Note that there is a different perturbation $\bdelta$ optimized for each sampled $\bI$ from the dataset $\mathcal{D}_{c}$. 



\noindent\textbf{Temporal Matching Learning in VOT}. Suppose the clean video training dataset consists of $n$ clean training videos, i.e., $\mathcal{D}_{v} = \{(V_{i}, B_{i})\}_{i=1}^{n}$,
where $V_{i}=\{\bI_{j}\}_{j=1}^{k}$ denotes the $i$-th clean video containing $k$ video frames, and $B_{i}=\{\bb_{j}\}_{j=1}^{k}$ represents the corresponding box annotations. In each frame, the annotation $\bb_{j} \in \real^{4}$ specifies the top-left coordinates, width, and height of the target. Typically, the deep tracker SiamFC \cite{SiamFC} is trained to perform template matching on randomly sampled pairs of frames within a training video:
\begin{equation}
\label{siamfc_label}
\min_{\theta}{\mathbb{E}}_{(\bz, \bx)\sim\mathcal{D}_{v}}{\mathcal{L}}(f_{\theta}(\bz) * f_{\theta}(\bx), y),
\end{equation}
where $f_{\theta}$ is the backbone in SiamFC parameterized with $\theta$. $\bz$ and $\bx$ are template and search images cropped in two randomly sampled frames $\bI_{i}$ and $\bI_{j}$ of a training video $V$. $*$ is the cross-correlation operation, and $y$ is the ground-truth response map indicating where the target is in $\bx$. $\mathcal{L}(\cdot,\cdot)$ represents the binary-cross entropy loss.

\begin{figure*}[t]
  \centering
  \includegraphics[width=0.9\linewidth]{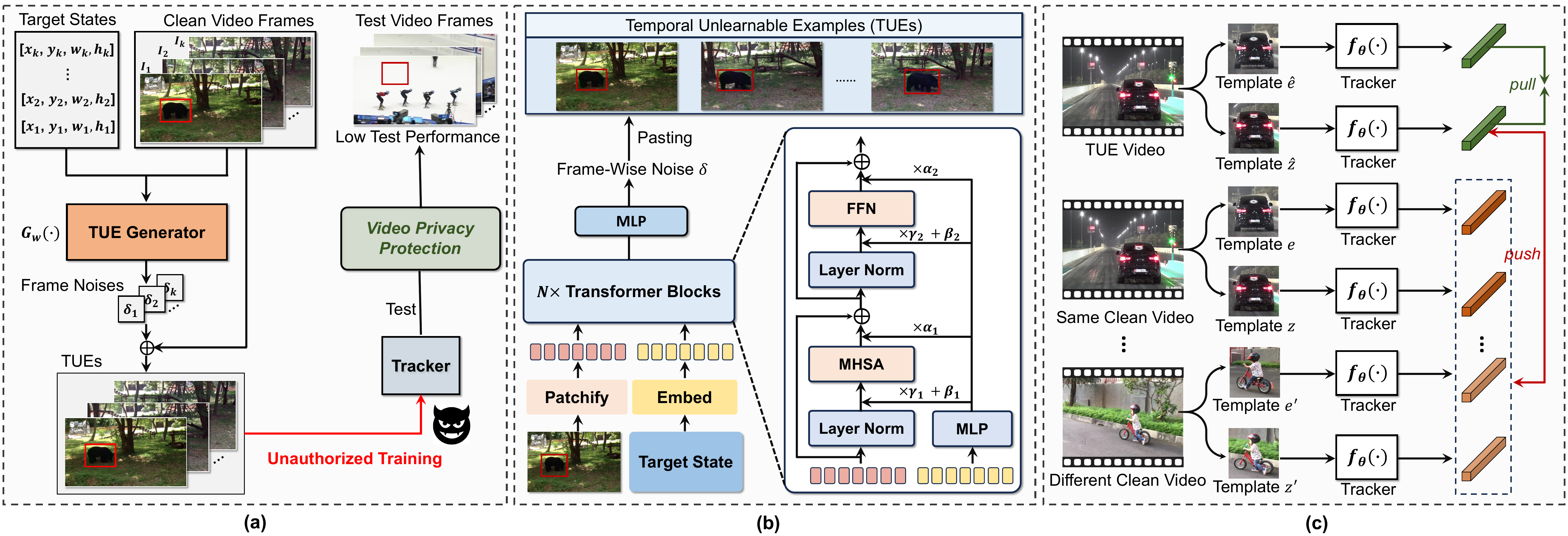}
  \vspace{-4mm}
   \caption{(a) Depiction of our
TUEs; (b) Architecture of the TUE generator; (c) The temporal contrastive learning scheme.}
   \label{overall_arch}
   \vspace{-4mm}
\end{figure*}

\subsection{Baselines: UEs for VOT}\label{sec:baseline}

Despite the progress of UEs for image classification, applying UEs to the VOT task is unexplored. 
To bridge this gap, we build new baselines by applying off-the-shelf EM \cite{em} to VOT \cite{SiamFC}. Specifically, the goal is to create perturbations that reduce the tracking loss function, thus exploiting a ``shortcut'' matching within the tracker: 
\begin{align}
\label{tue}
\begin{split}
&\quad\quad\min_{\theta}{\mathbb{E}}_{(\bz, \bx)\sim\mathcal{D}_{v}}\big[\min_{\bdelta_{t} \in \bDelta}\mathcal{L}(f_{\theta}(\hat{\bz})*f_{\theta}(\hat{\bx}),y)\big], \\[-4pt]
&\text{s.t.}~\hat{\bz}=\Phi(\bz,\phi(\bdelta_{t},\bb_{i}), \bb_{i}),~\hat{\bx}=\Phi(\bx,\phi(\bdelta_{t},\bb_{j}), \bb_{j}),
\end{split}
\end{align}
where $\hat{\bz}$ and $\hat{\bx}$ are the TUEs for the template and search images. 
The TUEs are created by interpolating the target perturbation $\bdelta_{t}$ to the same size as the bounding box, and pasting it onto the template/search image. 
Specifically, 
$\phi(\bdelta_t, \bb)$ is the Bilinear interpolation function, which interpolates $\bdelta_{t}$ to the same size as bounding box $\bb$, and $\Phi(\bx, \hat{\bdelta}_t, \bb)$ is the pasting function that pastes $\hat{\bdelta}_t$ onto the target region $\bb$ of image $\bx$ via the additional operation.
Since each video only contains one tracked instance, a single $\bdelta_{t}$ is defined for each video, \textit{i.e.,} shared across frames within the same video. 

In (\ref{tue}), the outer optimization updates the tracker  $\theta$, while the inner optimization optimizes the noise $\bdelta_{t}$ with the tracker fixed. These steps alternate during training, with the perturbation noise $\bdelta_{t}$ updated iteratively for $T$ steps using the PGD method \cite{pgd}. The above learning of TUEs $\hat{\bz}$ and $\hat{\bx}$ causes the tracker to rely on the perturbation noise for temporal matching, making the training videos unexploitable. Finally, the optimized noise set $\{\bdelta_{t}^{i}\}_{i=1}^{n}$ is obtained for each of the $n$ videos in $\mathcal{D}_v$.

\noindent\textbf{Context-Aware UEs.} 
Existing trackers \cite{SiamFC,ostrack,dropmae} incorporate both the central target and surrounding context regions as the template $\bz$ to enhance tracking performance. Inspired by this, we also take context regions into considerations when generating UEs to achieve more effective data privacy protection. Our approach follows the same aforementioned learning procedure, but now introduces context noise $\bdelta_{c}$ during optimization. More details are provided in the Supplementary.

\noindent\textbf{Tracker Training w/ UEs.} After obtaining the target and context perturbations $\{\bdelta_{t}^{i},\bdelta_{c}^{i}\}_{i=1}^{n}$, we interpolate the target noise $\bdelta_{t}^{i}$ and context noise $\bdelta_{c}^{i}$ to match the target and context sizes in each raw frame of a given video $V_{i}$. These noises are then pasted onto the corresponding regions in every frame, resulting in a new unlearnable dataset $\mathcal{D}_{u}$, which is used to train standard trackers \cite{ostrack,dropmae,chen2023seqtrack} following the official training protocols.

\subsection{Generative Framework for TUE}\label{sec:gen_framework}
The above baselines are based on the image-based 
approach EM, 
which has several limitations when applied to videos: 
1) the iterative optimization of $\hat{\bz}$ and $\hat{\bx}$ is time-consuming, each optimization needs $T$ iteration; 2) the noises $\bdelta_{t}$ and $\bdelta_{c}$ are pre-defined with fixed shapes, which need to be interpolated into various target scales in each frame of a video; 3) the generated noises $\{\bdelta_{t}^{i},\bdelta_{c}^{i}\}_{i=1}^{n}$ are video specific, which cannot generalize to other unseen video datasets.

To address the above limitations, in contrast to image-based methods, where the UEs are directly optimized, we propose a new generative framework for our TUEs generation. Fig. \ref{overall_arch} (a) depicts the overall pipeline of the proposed TUE generation process. Our TUE generation is formulated as follows:
\begin{equation}
\label{tue_context_gen}
\begin{split}
&\quad\quad~~\min_{\theta}{\mathbb{E}}_{(\bz, \bx)\sim\mathcal{D}_{v}}\big[\min_{{\mathbf{w}}}\mathcal{L}(f_{\theta}(\hat{\bz})*f_{\theta}(\hat{\bx}),y)\big], \\[-2pt]
& \!\!\text{s.t.}~\hat{\bz}\!=\!\bz \!+\! G_{\mathbf{w}}(\bz,\widetilde{\bb}_{i}),~\hat{\bx}\!=\!\Phi(\bx, G_{\mathbf{w}}(c(\bx,{\bb_{j}}), \widetilde{\bb}_{j}), \bb_{j}),
\end{split}
\end{equation}
where $\hat{\bz}$ and $\hat{\bx}$ are the generated TUEs for the template and search images.
These TUEs are obtained by generating target-aware perturbation noises, and then pasting them onto the corresponding target regions in video frames.
Specifically,  $G_{\mathbf{w}}(\cdot)$ is our TUEs generator parameterized with $\mathbf{w}$, which takes a target patch $\bz$ and normalized bounding box $\widetilde{\bb}_i$ as input, and generates a noise perturbation $G_{\mathbf{w}}(\bz,\widetilde{\bb}_{i})$ that is added back to $\bz$.
For the search image $\bx$, the target is cropped using the cropping function $c(\bx,\bb_j)$ via the given box annotation $\bb_{j}$, and then passed to the TUE generator with its corresponding normalized bounding box $\widetilde{\bb}_j$ to generate the perturbation $G_{\mathbf{w}}(c(\bx,{\bb_{j}}), \widetilde{\bb}_{j})$. This perturbation is then pasted back onto the target in the search image via the $\Phi(\cdot)$ pasting function. 
Note that $\widetilde{\bb}_i, \widetilde{\bb}_j \in \real^{4}$ are the normalized target states (\textit{i.e.,} containing the normalized top-left coordinates, width and height) of the corresponding $\bz$ and $c(\bx,{\bb_{j}})$ patches, respectively. 


We employ a diffusion architecture, \textit{i.e.,} DiT \cite{dit} as the TUE generator, which takes 
class label and time step as the condition for multi-step 
image generation. Here we adapt it to generate the target-aware TUEs in a single feed-forward step with the normalized target condition, which is more efficient. The architecture is in Fig. \ref{overall_arch} (b). 


\begin{algorithm}[t]
\caption{Optimization of TUE generator}
\SetKwInOut{Input}{Input}\SetKwInOut{Output}{Output}
\Input{Surrogate model $f_{\mathbf{\theta}}$, TUE generator $G_{\mathbf{w}}$, learning rates $\alpha_s$ and $\alpha_g$, number of epochs $ep$, clean dataset $\mathcal{D}_{v} = \{(V_{i}, B_{i})\}_{i=1}^{n}$}
\Output{Optimized TUE generator $G_{\mathbf{w}}$}
\SetKwData{ModelParam}{ model.parameters}
\SetKwFunction{Clip}{ { Clip}}
\SetKwFunction{GradMod}{ {ReGrad}}
\SetKwFunction{trainable}{ {IsTrainable}}

\For{$i \leftarrow 1$ \textbf{to} $ep$ }{
\For{$(\bz, \bx) \in \mathcal D_{v}$}{
\textcolor{teal}{\# Generate perturbations}\\
$\hat{\bz}\!=\!\bz \!+\! G_{\mathbf{w}}(\bz,\widetilde{\bb}_{i})$,\\
$\hat{\bx}\!=\!\Phi_{c}(\bx,G_{\mathbf{w}}(c(\bx,{\bb_{j}}),\widetilde{\bb}_{j}),\bb_{j})$;\\
\textcolor{teal}{\# Optimize perturbators $G_{\mathbf{w}}$ using $\alpha_s$}\\
$\hat{\mathbf{e}}=c(\bx,{\bb_{j}})+G_{\mathbf{w}}(c(\bx,{\bb_{j}}),\widetilde{\bb}_{j})$,\\
$\mathcal{L}_{f}\!=\!\mathcal{L}(f_{\mathbf{\theta}}(\hat{\bz})\!*\!f_{\mathbf{\theta}}(\hat{\bx}),y)\!+\!\lambda \mathcal{L}_{cl}(f_{\mathbf{\theta}}(\hat{\bz}),f_{\mathbf{\theta}}(\hat{\mathbf{e}}))$,\\
Optimize $G_{\mathbf{w}}$ via Adam to minimize $\mathcal{L}_f$;\\
\textcolor{teal}{\# Optimize the surrogate model $f_{\mathbf{\theta}}$ using $\alpha_g$} \\
$\mathcal{L}_{f}\!=\!\mathcal{L}(f_{\mathbf{\theta}}(\hat{\bz})\!*\!f_{\mathbf{\theta}}(\hat{\bx}),y)$,\\
Optimize $f_{\mathbf{\theta}}$ via Adam to minimize $\mathcal{L}_f$;
}
}
\label{alg:train}
\end{algorithm}

\textbf{\uline{Advantages:}} 
Our proposed generative TUE framework is more efficient than the EM baseline, in both training scalability and inference, and also transfers well to unseen videos. Specifically: 
1) \textbf{High Training efficiency}: since $G_{\mathbf{w}}(\cdot)$ is lightweight and can be directly updated via the gradient back-propagation in each inner optimization, the training of $G_{\mathbf{w}}(\cdot)$ is efficient, e.g., $4\times$ faster than the EM baseline as illustrated in Table~\ref{complexicity_analysis}; 
2) \textbf{Fewer learnable parameters}: The EM baseline optimizes both target and context noises for each video (size 127$\times$127$\times$3), resulting in a large number of parameters on popular tracking benchmarks (e.g., 3.4GB for GOT-10k \cite{got10k}). In contrast, our approach optimizes a lightweight model $G_{\mathbf{w}}(\cdot)$ with a fixed number of parameters (124MB), which is independent of dataset size;
3) \textbf{Applicable to Unseen Videos}: The EM baseline requires optimizing UEs for each dataset, which means it needs to be retrained to generate UEs for unseen datasets. As shown in Table \ref{dataset_transfer}, our TUE approach can generate UEs for unseen datasets directly through model inference, without the need for retraining.

\begin{figure}[t]
  \centering
  \includegraphics[width=0.9\linewidth]{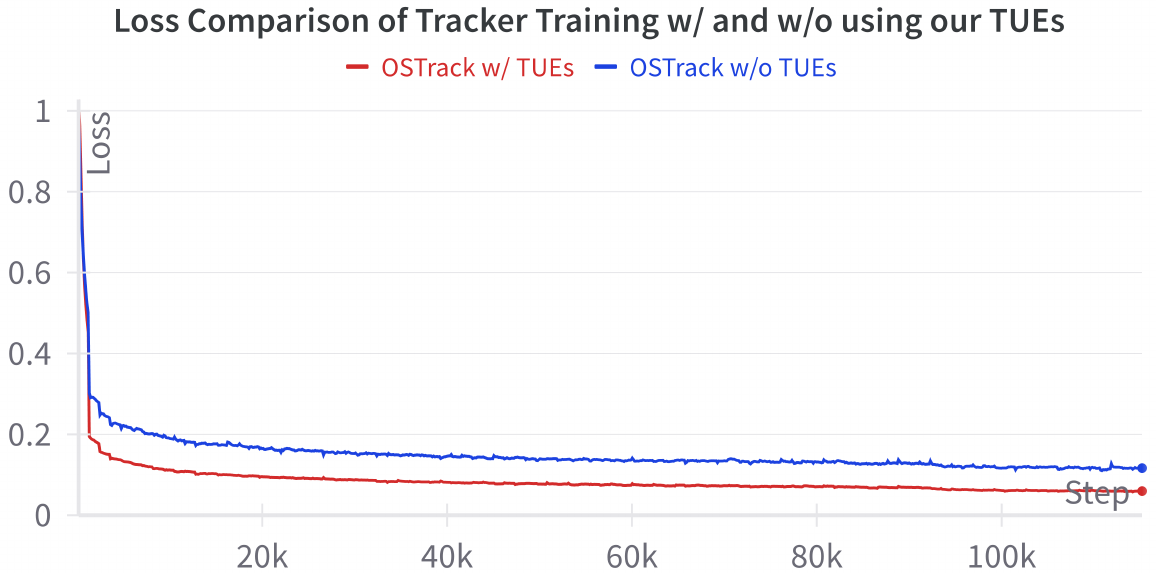}
  \vspace{-3mm}
   \caption{Training loss of OSTracker w/ and w/o our TUEs. }
   \label{loss_comp}
   \vspace{-4mm}
\end{figure}

\subsection{Temporal Contrastive Learning}\label{sec:tcl}
We further propose a Temporal Contrastive Loss (TCL) to encourage the tracker to rely more on the generated TUEs for temporal matching, thereby further degrading the model's tracking performance on clean test videos.
Fig. \ref{overall_arch} (c) illustrates the details of the proposed Temporal Contrastive Learning scheme. 
Specifically, the template TUE
$\hat{\bz}$ is used as the exemplar, treating the TUE $\hat{\mathbf{e}}$ (in the search region $\bx$) within the same video as the positive sample. The clean templates in the same video (\textit{i.e.,} $\bz$ and $\mathbf{e}=c(\bx,{\bb_{j}})$) and the other videos (\textit{i.e.,} $\bz^{\prime}$ and $\mathbf{e}^{\prime}$) within the same batch are regarded as the negative samples to $\hat{\bz}$.
The resulting formulation is:
\vspace{-2mm}
\begin{equation}
\label{tue_context_gen_cl}
\begin{split}
&\min_{\mathbf{w}}[\mathcal{L}(f_{\theta}(\hat{\bz})*f_{\theta}(\hat{\bx}),y)+\lambda \mathcal{L}_{cl}(f_{\theta}(\hat{\bz}),f_{\theta}(\hat{\mathbf{e}}))],\\[-2pt]
&\text{s.t.} \quad\hat{\mathbf{e}}=c(\bx,{\bb_{j}})+G_{\mathbf{w}}(c(\bx,{\bb_{j}}),\widetilde{\bb}_{j}), 
\end{split}
\end{equation}

\vspace{-3mm}
\noindent where $\hat{\mathbf{e}}$ is the TUE in the searching region $\bx$, and $\mathcal{L}_{cl}(\cdot)$ is the contrastive loss \cite{simclr}. Note that we use the above objective to perform the inner optimization, while employing the tracker loss $\mathcal{L}(\cdot)$ for outer optimization to keep consistent with the original tracker training process. 
The overall optimization process of the TUE generator is outlined in Alg.~\ref{alg:train}. In each iteration, we first optimize the generator, followed by the surrogate model optimization. {More details on the proposed temporal contrastive learning are in the supplementary.}

\noindent\textbf{Tracker training w/ TUEs.} With the learned generator $G_{\mathbf{w}}(\cdot)$, we generate perturbations for existing tracking datasets and apply them to bounding boxes to create TUEs. These TUEs are then used to train various trackers following their official settings. As shown in Fig.~\ref{loss_comp}, OSTrack trained with TUEs exhibits lower training loss, as it learns a ``shortcut," relying on perturbations for temporal matching while ignoring the original data structure, thereby preserving training privacy.

\subsection{Practical Application Pipeline}\label{sec:application}
{Fig.~\ref{workflow} illustrates our four-stage application pipeline. In Stage 1, we train the TUE generator $G_{\mathbf{w}}$ using the surrogate tracker SiamFC on an off-the-shelf tracking dataset (e.g., GOT-10k) under the supervision of Eq.~\ref{tue_context_gen_cl}. In Stage 2, the trained $G_{\mathbf{w}}$ generates TUEs to protect user video data. Stage 3 involves training an unauthorized tracker on the protected data. Finally, in Stage 4, the trained tracker is deployed on clean data to evaluate privacy protection performance. All experimental results are derived from Stage 4. We conduct experiments across multiple tracking applications, including VOT, VOS, and long-term point tracking, demonstrating that the protected videos effectively prevent unauthorized exploitation by these applications.}


\begin{figure}[t]
  \centering
  \includegraphics[width=0.85\linewidth]{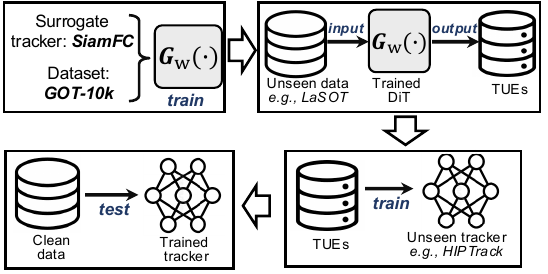}
  \vspace{-3mm}
   \caption{Overall workflow of our proposed TUEs.}
   \label{workflow}
   \vspace{-5mm}
\end{figure}

\section{Experiments}
\subsection{Implementation Details}

Our TUE generator $G_{\mathbf{w}}(\cdot)$ is implemented as a lightweight DiT-S/8 model \cite{dit} with 12 layers, 6 attention heads, and a hidden state size of 384. A fully connected layer maps the target state to the hidden space for controllable TUE generation.  
%
The generator is jointly trained with the naive SiamFC tracker \cite{SiamFC} using Adam \cite{Adam} for 50 epochs with a learning rate of $5\times10^{-6}$ and a batch size of 16 (Algorithm~\ref{alg:train}). We set $\lambda$ to 0.05. Following \cite{SiamFC}, the generator processes cropped template patches for efficiency. GOT-10k \cite{got10k} is used as the source training dataset. 
SiamFC is chosen as the base tracker for two reasons: 1) its efficient training enables effective learning of the TUE generator, requiring only 7 hours on a single NVIDIA 4090 GPU; 2) the trained generator generalizes well to more complex trackers and datasets, avoiding time-intensive optimization with larger models. 
Once trained, the generator creates TUEs offline for training various trackers.

\noindent\textbf{Evaluation.} The trained models are evaluated on widely used tracking benchmarks, including GOT-10k \cite{got10k}, LaSOT \cite{lasot}, OTB \cite{OTB100}, DAVIS-17 \cite{davis17}, and YTVOS-19 \cite{youtubevos}, using their standard evaluation metrics. Lower performance indicates stronger privacy protection.

\subsection{Ablation Studies}

\noindent\textbf{The effect of TCL.}
As illustrated in Tab. \ref{compo_ana}, using our TCL leads to larger performance degradation on both SiamFC and OSTrack. This is because TCL leads to larger distribution gap between clean samples and TUEs, which hinders the trackers' generalization to clean test data, thus ensuring training data privacy.

\begin{table}[t]
\centering
\scriptsize
\resizebox{0.475\textwidth}{!}{ 
\begin{tabular}{@{}c|c|cc|ccc@{}}
\toprule
\multirow{2}{*}{Trackers} & \multirow{2}{*}{Variants} & \multicolumn{2}{c|}{OTB \cite{OTB100}} & \multicolumn{3}{c}{GOT-10k \cite{got10k}} \\
& & AUC & Prec. & AO & SR$_{0.5}$ & SR$_{0.75}$ \\
\midrule
\multirow{6}{*}{SiamFC} & Clean & 58.6 & 79.2 & 35.5 & 39.0 & 11.8 \\
\cmidrule{2-7}
& EM \cite{em} Baseline & 39.6 & 54.7 & 27.0 & 25.7 & 5.8 \\
& \quad +Context & 29.5 & 37.4 & 21.4 & 18.2 & 4.3 \\
\cmidrule{2-7}
& TUE Generator & 17.6 & 19.9 & 16.1 & 10.4 & \textbf{1.5} \\
& \quad - Condi. & 19.7 & 22.4 & 22.5 &19.4&4.6 \\
& \quad + TCL & \textbf{11.4} & \textbf{13.5} & \textbf{12.1} & \textbf{9.0} & 1.9 \\
\hline
\multirow{3}{*}{OSTrack} & Clean & 67.4 & 89.4 & 71.0 & 80.4 & 68.2 \\
\cmidrule{2-7}
& TUE Generator & 45.3 & 60.6 & 41.6 & 47.3 & 30.3 \\
& \quad + TCL & 30.5 & 45.8 & 18.0 & 15.1 & 4.6 \\
\bottomrule
\end{tabular}
} 
\vspace{-0.3cm}
\caption{Results of SiamFC and OSTrack trained on GOT-10k with clean videos (Clean); videos modified with EM baseline and further enhanced by including context optimiziation (``+Context''); and videos modified with our proposed TUE (``Ours''), and with temporal contrastive learning (``+TCL''). 
}
\label{compo_ana}
\vspace{-4mm}
\end{table}

\begin{figure}[t]
  \centering
  \includegraphics[width=0.85\linewidth]{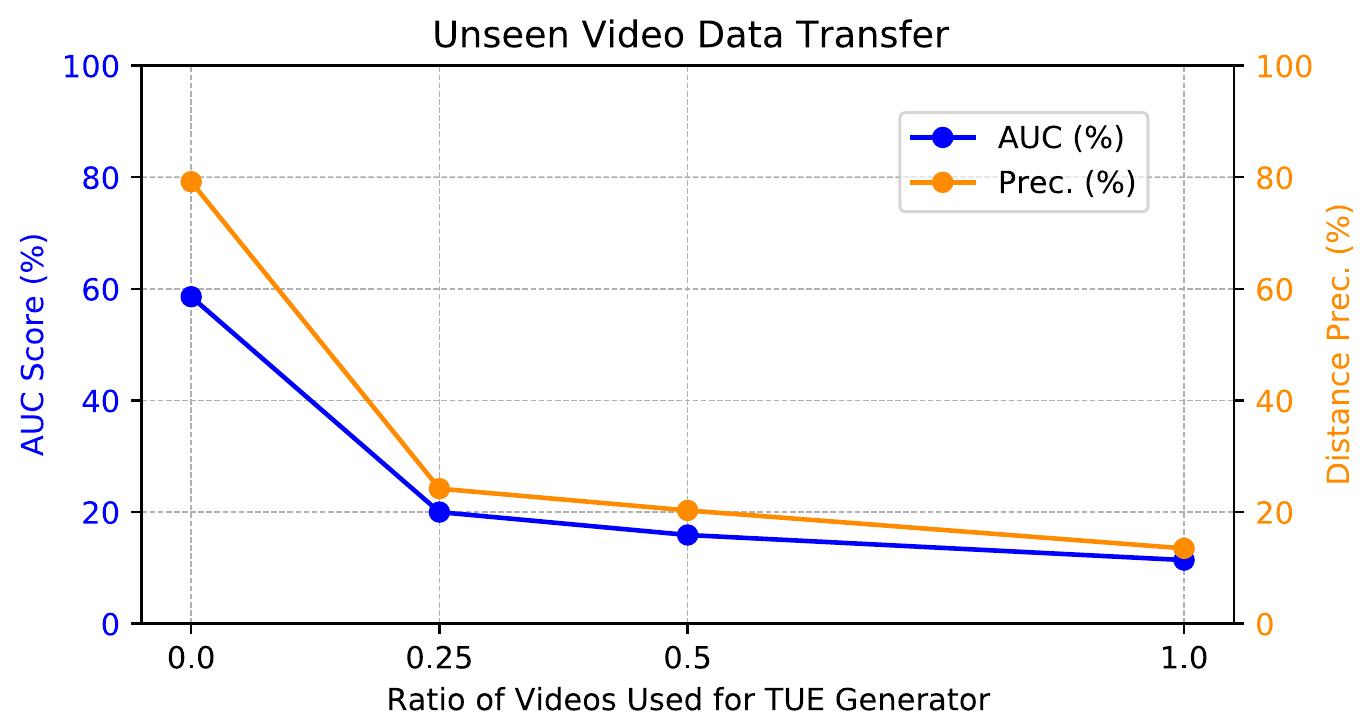}
  \vspace{-4mm}
   \caption{Ablation study on the data ratio in GOT-10k \cite{got10k} used for training the TUE generator.}
   \vspace{-5mm}
   \label{data_ratio}
\end{figure}

\noindent\textbf{Incorporating context noise for optimization.}
In Tab. \ref{compo_ana}, we find that our EM baseline that only optimizes the target perturbation noise has limited effect. %
Incorporating the context noise for optimization leads to larger performance drops on both OTB and GOT-10k testing datasets. This indicates corrupting the target region only is not effective enough for SiamFC, i.e., the trackers can learn temporal matching from the context regions.

\noindent\textbf{The usage of target state condition.} In Eq. \ref{tue_context_gen}, we remove the target stage condition $\hat{\bb}_{i}$ (``- Condi.'') and only use the input image for TUE generation. This variant ignores the target state,  achieving inferior protection performance, which shows that dynamically adapting to the target state is helpful for TUE learning.

\noindent\textbf{Data ratio for training TUE generator.} As shown in Fig. \ref{data_ratio}, we train our TUE generator using different video ratios from GOT-10k \cite{got10k}. Specifically, we train the generator on a subset of randomly selected videos and then use it to generate TUEs for the entire dataset, which are used to fine-tune a SiamFC tracker. Results show that the TUE generator can be effectively learned with just 25\% of the GOT-10k videos (about 2,300 videos) and generate effective TUEs on unseen videos.


\noindent\textbf{Model complexity.} As shown in Tab. \ref{complexicity_analysis}, training the TUE generator takes 7 hours on a single RTX4090, significantly faster than the EM-based iterative optimization. Additionally, our approach uses fewer learnable model parameters, leading to more efficient training.


\begin{table}[t]
\centering
\footnotesize
\resizebox{0.45\textwidth}{!}{ 
\begin{tabular}{ccc}
\toprule
Method & Training Time & Learnable Parameter Size \\
\midrule
EM + Context & 33 Hours  & 3.4GB \\
TUE Generator & 7 Hours &124MB \\
\bottomrule
\end{tabular}
} 
\vspace{-0.3cm}
\caption{Model complexity of our TUE and EM baseline. }
\label{complexicity_analysis}
\vspace{-4mm}
\end{table}

\begin{table}[t]
\centering
\scriptsize
\resizebox{0.48\textwidth}{!}{ 
\begin{tabular}{c|c|ccc}
\toprule
 \multirow{2}{*}{Variants} & SeqTrack  & \multicolumn{3}{c}{GOT-10k} \\
& Train. Epoch & AO & SR$_{0.5}$ & SR$_{0.75}$ \\
\midrule
Clean & 500 & 74.7 &84.7 &71.8 \\  
\midrule
TUE Generator&100 & 9.7 (65.0 $\downarrow$) &4.1 (80.6 $\downarrow$)&0.5 (71.3 $\downarrow$)\\
TUE Generator& 200&\bf{ 2.1} (72.6 $\downarrow$)&0.9 (83.8 $\downarrow$)&0.1 (71.7 $\downarrow$) \\
TUE Generator& 300 & 2.2 (72.5 $\downarrow$)&1.2 (83.5 $\downarrow$)&0.2 (71.6 $\downarrow$) \\
TUE Generator& 400 & 3.8 (70.9 $\downarrow$)&1.2 (83.5 $\downarrow$)&0.1 (71.7 $\downarrow$) \\
TUE Generator& 500 & 2.3 (72.4 $\downarrow$)&\bf{0.7} (\bf{84.0} $\downarrow$)&\bf{0.0} (\bf{71.8} $\downarrow$) \\
\bottomrule
\end{tabular}
} 
\vspace{-0.3cm}
\caption{Training SeqTrack \cite{chen2023seqtrack} with TUEs generated by our TUE generator on GOT-10k for various training epochs.  }
\label{train_epoch}
\vspace{-4mm}
\end{table}

\begin{table*}[t]
\scriptsize
\centering
 \setlength\tabcolsep{4.5pt}
\scalebox{1.0}{
\begin{tabular}{@{}c|c|ccc|cc|ccc@{}}
\toprule
\multirow{2}{*}{\begin{tabular}[c]{@{}c@{}}VOT Method\end{tabular}} &  \multirow{2}{*}{\begin{tabular}[c]{@{}c@{}}UE Method\end{tabular}}& \multicolumn{3}{c|}{GOT-10k \cite{got10k}}& \multicolumn{2}{c|}{OTB-100 \cite{OTB100}}  & \multicolumn{3}{c}{LaSOT \cite{lasot}}                         \\
& & AO & SR$_{0.5}$ & SR$_{0.75}$ & AUC & P & AUC & P$_{Norm}$ & P \\

\midrule
\multirow{6}{*}{SiamFC \cite{SiamFC}}     &    Clean     &35.5 &	39.0 &	11.8  & 58.6 & 79.2 & 34.0 & 39.9 & 33.0    \\
& TAP \cite{tap}&32.9&35.0&9.5&56.7&76.5&31.4&36.7&29.8\\
& LSP \cite{lsp}&28.1&29.3&7.9&50.0&67.9&26.6&30.9&24.2 \\
& AR \cite{ar}&34.1&37.2&11.5&56.9&76.7&33.6&38.9&32.3\\
& EM \cite{em}                                                           & 		21.4 &	18.2 &	4.3 &  	29.5 &	37.4 & 17.6 & 19.6 & 14.5           \\
&  \textbf{TUE (Ours)} & \textbf{12.1 (23.4$\downarrow$)} &	\textbf{9.0 (30.0$\downarrow$)} &  \textbf{1.9 (9.9$\downarrow$)} & \textbf{11.4 (47.2$\downarrow$)} & \textbf{13.5 (65.7$\downarrow$)} & \textbf{9.5 (24.5$\downarrow$)} & \textbf{9.5 (30.4$\downarrow$)} & \textbf{6.4 (26.6$\downarrow$)}        \\
\midrule
\multirow{6}{*}{OSTrack-256 \cite{ostrack}}  & Clean &71.0 &	80.4 &	68.2  & 67.4 &	89.4 & 62.3 & 70.2 & 66.0 \\
& TAP \cite{tap}& \textbf{17.0}&17.3&10.2&44.9&61.7&26.8&32.2&29.1  \\
& LSP \cite{lsp}  & 27.8&29.7&16.4&48.2&66.1&28.3&33.3&30.2 \\
& AR \cite{ar}&67.0&75.9&62.4&63.8&85.1&59.2&67.0&62.0 \\
& EM \cite{em} & 26.3 &	24.0 &10.2   & 48.8 &	69.6  & 29.9 & 36.2 & 31.8              \\
& \textbf{TUE (Ours)} & 18.0 (53.0$\downarrow$) &	\textbf{15.1 (65.3$\downarrow$)} &	\textbf{4.6 (63.6$\downarrow$)} & \textbf{30.5 (36.9$\downarrow$)} &	\textbf{45.8 (43.6$\downarrow$)} & \textbf{22.0 (40.3$\downarrow$)} & \textbf{27.7 (42.5$\downarrow$)} & \textbf{23.1 (42.9$\downarrow$)}  \\
\midrule
\multirow{6}{*}{DropTrack-384 \cite{dropmae}}     &    Clean  & 75.9 &	86.8 & 72.0 & 69.4 & 91.3 & 66.5 & 75.2 & 71.5 \\
& TAP \cite{tap}&51.7&57.9&41.8&51.4&70.6&39.2&45.9&40.4\\
& LSP \cite{lsp}&27.8&29.3&20.3&66.5&87.7&38.3&44.4&39.8 \\
& AR \cite{ar}&66.9&75.5&61.9&66.4&87.6&60.4&68.4&63.7\\
& EM \cite{em}  & 21.6 &	18.7 &	7.0 & 48.7 &	72.2 & 33.3 & 40.9 & 36.4 \\
& \textbf{TUE (Ours)} & \textbf{17.1 (58.8$\downarrow$)} &	\textbf{12.9 (73.9$\downarrow$)} &	\textbf{2.7 (69.3$\downarrow$)} & \textbf{36.7 (32.7$\downarrow$)} & \textbf{57.7 (33.6$\downarrow$)}  & \textbf{25.2 (41.3$\downarrow$)} & \textbf{32.2 (43.0$\downarrow$)} & \textbf{28.3 (43.2$\downarrow$)} \\
\midrule
\multirow{6}{*}{SeqTrack-256 \cite{chen2023seqtrack}} &  Clean  & 74.7 &	84.7 &	71.8 & 68.1 & 89.9 & 63.6 & 72.4 & 67.9 \\
& TAP \cite{tap}& 8.1&8.6&4.4&39.4&57.0&18.1&23.1&21.6\\
& LSP \cite{lsp}  &13.5&14.3&6.8&44.5&63.8&25.0&31.4&28.1 \\
& AR \cite{ar}&64.3&73.1&59.1&64.5&85.7&56.0&64.3&58.7\\
& EM \cite{em} & 8.1 &	6.0 & 1.1 & 34.7 & 55.2 & 15.3 & 21.3 & 19.9 \\
& \textbf{TUE (Ours)} & \textbf{2.1 (72.6$\downarrow$)} &	\textbf{0.9 (83.8$\downarrow$)} & \textbf{0.1 (71.7$\downarrow$)} & \textbf{7.0 (61.1$\downarrow$)} & \textbf{14.3 (75.6$\downarrow$)} & \textbf{3.4 (60.2$\downarrow$)} & \textbf{5.4 (67.0$\downarrow$)} & \textbf{6.3 (61.6$\downarrow$)} \\
\midrule
\multirow{6}{*}{MixFormer-CvT \cite{MixFormer}} & Clean & 70.7 &80.0 &	67.8 & 66.1 & 88.6 & 62.1 & 69.9 & 65.6 \\
& TAP \cite{tap}&12.4&11.0&4.6&40.4&53.9&27.7&31.8&27.7\\
& LSP \cite{lsp}&20.9&21.9&10.9&44.7&60.8&31.2&36.0&31.2 \\
& AR \cite{ar}&51.4&57.9&43.9&62.0&81.6&57.2&64.7&58.6\\
& EM \cite{em}  & 7.0 &	6.8 & 	2.7  & 45.1 &	62.0 & 24.5 & 30.6 & 27.7 \\
& \textbf{TUE (Ours)} & \textbf{1.9 (68.8$\downarrow$)} & \textbf{0.1 (79.9$\downarrow$)} & \textbf{0.2 (67.6$\downarrow$)} & \textbf{14.7 (51.4$\downarrow$)} & \textbf{22.9 (65.7$\downarrow$)} & \textbf{8.1 (54.0$\downarrow$)} & \textbf{10.7 (59.2$\downarrow$)} & \textbf{11.4 (54.2$\downarrow$)} \\
\midrule
\multirow{6}{*}{STARK-S50 \cite{stark}} & Clean & 67.2 & 76.1 & 	61.2 & 64.1 & 84.7 & 58.2 & 65.7 & 59.5    \\
& TAP \cite{tap}&18.6&15.6&6.6&18.7&21.7&18.3&10.8&6.0\\
& LSP \cite{lsp}&8.7&5.0&1.5&22.7&29.4&11.4&\textbf{8.1}&\textbf{2.2} \\
& AR \cite{ar}&51.7&51.9&41.8&52.9&69.5&46.7&44.6&35.5\\
& EM \cite{em} & 14.8 & 15.1 &	8.3 &  43.4 &	56.3 & 28.3 & 31.3 & 26.7\\
& \textbf{TUE (Ours)} & \textbf{2.6 (64.6$\downarrow$)} & \textbf{1.1 (75.0$\downarrow$)} & \textbf{0.2 (61.0$\downarrow$)} &  \textbf{13.9 (50.2$\downarrow$)} & \textbf{17.2 (67.5$\downarrow$)} & \textbf{8.9 (49.3$\downarrow$)} & 8.9 (56.8$\downarrow$) & 7.3 (52.2$\downarrow$) \\
\midrule
\multirow{3}{*}{AQATrack-256 \cite{aqatrack}}     &   Clean &73.2&82.6&71.5& 69.1 & 91.5 & 64.3 & 72.7 & 69.0                                                                \\
& EM \cite{em} & 19.4	&16.2 &	4.9 & 47.1 & 65.3 & 31.1 & 37.2 & 31.8 \\
& \textbf{TUE (Ours)} & \textbf{16.2 (57.0$\downarrow$)} &	\textbf{10.6 (72.0$\downarrow$)} &	\textbf{1.7 (69.8$\downarrow$)} & 	\textbf{20.8 (48.3$\downarrow$)} &	\textbf{28.9 (62.6$\downarrow$)} & \textbf{17.4 (46.9$\downarrow$)} & \textbf{20.3 (52.4$\downarrow$)} & \textbf{16.7 (52.3$\downarrow$)}          \\
\midrule
\multirow{3}{*}{HIPTrack \cite{hiptrack}}     &    Clean  & 77.4 & 88.0 & 74.5 & 68.8  & 90.3 & 66.8 & 75.1 & 72.0    \\
& EM \cite{em}                                                                        & 63.0 &	71.9 & 52.7  & 66.9 & 88.9 & 55.5 & 64.0 & 59.3           \\
& \textbf{TUE (Ours)}                                                                    & \textbf{43.2 (34.2$\downarrow$)} &	\textbf{43.9 (44.1$\downarrow$)} &	\textbf{20.0 (54.5$\downarrow$)} & \textbf{56.9 (11.9$\downarrow$)} & \textbf{78.6 (11.7$\downarrow$)} & \textbf{38.7 (28.1$\downarrow$)} & \textbf{48.5 (26.6$\downarrow$)} & \textbf{42.7 (29.3$\downarrow$)} \\
\bottomrule
\end{tabular} 
}
  \vspace{-0.3cm}
\caption{Methodology transfer on various trackers. We use TUEs, which are specifically optimized with SiamFC on the GOT-10k training set, to train SOTA deep trackers. The trained trackers are tested on clean GOT-10k, OTB-100, and LaSOT test sets. Performance drops of our method are shown in brackets. The best results are shown in bold.}
\label{overall_results}
 \vspace{-1mm}
\end{table*}

\begin{table*}[t]
\centering
\scriptsize
 \setlength\tabcolsep{5.5pt}
\scalebox{1.0}{
\begin{tabular}{c|c|ccc|cc|ccc}
\toprule
\multirow{2}{*}{\begin{tabular}[c]{@{}c@{}}Training Dataset\end{tabular}} &  \multirow{2}{*}{\begin{tabular}[c]{@{}c@{}}UE Method\end{tabular}}& \multicolumn{3}{c|}{GOT-10k \cite{got10k}}& \multicolumn{2}{c|}{OTB-100 \cite{tnl2k}}  & \multicolumn{3}{c}{LaSOT \cite{lasotext}}                         \\
& & AO & SR$_{0.5}$  & SR$_{0.75}$  & AUC  & P  & AUC & P$_{Norm}$ & P  \\

\midrule
\multirow{5}{*}{LaSOT} & Clean                                                                  & 55.6 &	62.1 & 44.6 &49.2 &	64.0 & 58.2 & 65.7 & 59.5        \\
  & AR \cite{ar}  & 52.3 & 57.9 & 40.1 & 45.3 & 58.5 &   52.4 & 59.6 & 53.3                                                                         \\
    & LSP \cite{lsp}   & 15.8&12.8&5.0& 23.5 & 31.0 & 18.1 & 20.3 & 17.0                                                                          \\
     & EM \cite{em} &  11.4 &9.9 & 7.9 & 21.9 & 29.8 & 18.9 & 22.0 & 19.9                                                                          \\
 & TUE (Ours)                                                                  & \textbf{4.1 (51.5$\downarrow$)}	& \textbf{3.0 (59.1$\downarrow$)} & \textbf{0.8 (43.8$\downarrow$)} & 	\textbf{9.5 (39.7$\downarrow$)} &	\textbf{15.7 (48.3$\downarrow$)} & \textbf{5.7 (52.5$\downarrow$)} & \textbf{7.6 (58.1$\downarrow$)} & \textbf{7.6 (51.9$\downarrow$)}          \\
 \midrule
 \multirow{5}{*}{LaSOT+GOT-10k} & Clean   & 66.2 &	76.1	& 59.7 & 64.6 & 85.4 & 62.1 & 70.9 & 65.1                                                                         \\
  & AR \cite{ar}   & 61.2 &70.0&53.9& 60.9 & 80.5 &  58.2 & 66.1 & 59.3                                                                        \\
   & LSP \cite{lsp}   & 27.0&25.8&11.6&36.6&47.5&30.2&31.6&25.9                                                                         \\
  & EM \cite{em} &20.6&	20.5&11.3&27.2&35.7&22.4&24.5&20.6                                                                           \\
 & TUE (Ours)                                                                  & \textbf{4.0 (62.2$\downarrow$)} & \textbf{2.1 (74.0$\downarrow$)} & \textbf{0.6 (59.1$\downarrow$)} & \textbf{11.0 (53.6$\downarrow$)} & \textbf{16.9 (68.5$\downarrow$)} & \textbf{6.8 (55.3$\downarrow$)} & \textbf{8.4 (62.5$\downarrow$)} & \textbf{7.8 (57.3$\downarrow$)}          \\
\bottomrule
\end{tabular} 
}
  \vspace{-0.3cm}
\caption{Unseen dataset transfer. We use our TUE generator, which is specifically optimized with SiamFC on the GOT-10k training set, to perform zero-shot TUEs generation on the unseen LaSOT training set. The obtained TUE-perturbed LaSOT and the different combination (i.e., TUE-perturbed LaSOT + GOT-10k) are used to train the base tracker STARK-S50 \cite{stark}. }
\label{dataset_transfer}
 \vspace{-4mm}
\end{table*}

\noindent\textbf{Training epochs vs.~performance drop.} In Tab. \ref{train_epoch}, we train SeqTrack-256 \cite{chen2023seqtrack} on TUE-GOT10k to investigate the impact of training epochs. 
Training for only 200 epochs with TUEs effectively degrades its performance. 
Note that SeqTrack-256 originally uses 500 epochs, which demonstrates that TUEs can effectively corrupt the tracker training w/ less training epochs. 

\noindent\textbf{Bbox dependency.} Our TUE generator requires the target bounding box as the input for target-aware TUEs generation. In practical applications, users can annotate short video clips manually or employ off-the-shelf trackers to generate reliable pseudo bboxes, similar to the annotations used in TrackingNet \cite{trackingnet}. 
In addition, we also validate that our TUEs can be automatically generated (using naive unsupervised EdgeBox to generate bbox proposals) in Tab. R3 of the Supplementary, eliminating the need for user intervention.

\subsection{Transfer Experiments}
%

Transferability is crucial in real-world applications \cite{yu2025mtlue, yu2025towards}. 
We evaluate our method's transferability to various trackers, datasets, and the dense temporal matching task.

\begin{table*}[t]
\centering
\scriptsize
\setlength\tabcolsep{5.5pt}
\scalebox{1.0}{
\begin{tabular}{c|c|c|ccc|ccc}
\toprule
\multirow{2}{*}{\begin{tabular}[c]{@{}c@{}}VOS Method\end{tabular}} & \multirow{2}{*}{\begin{tabular}[c]{@{}c@{}}Training Dataset\end{tabular}} &  \multirow{2}{*}{\begin{tabular}[c]{@{}c@{}}UE Method\end{tabular}}& \multicolumn{3}{c|}{DAVIS-17 Val \cite{davis17}} & \multicolumn{3}{c}{YTVOS-19 Val \cite{youtubevos}}                       \\

& &        & $\mathcal{J}\&\mathcal{F}$           & $\mathcal{J}$      & $\mathcal{F}$ & $\mathcal{J}\&\mathcal{F}$          & $\mathcal{J}_{seen}$ & $\mathcal{J}_{unseen}$\\
 \midrule
 \multirow{7}{*}{STCN \cite{STCN}} & \multirow{2}{*}{DAVIS-17} & Clean  &  71.2 & 67.3 & 72.7 & 63.3 & 66.3 & 56.1 \\
& & TUE (Ours)  &  \textbf{50.1 (21.1$\downarrow$)} & \textbf{46.8 (20.5$\downarrow$)} & \textbf{53.4 (19.3$\downarrow$)} & \textbf{42.4} (\textbf{20.9}$\downarrow$) & \textbf{42.7} (\textbf{23.6}$\downarrow$) & \textbf{36.9} (\textbf{19.2}$\downarrow$) \\
\cmidrule{2-9}
 &  \multirow{5}{*}{DAVIS-17+YTVOS-19} & Clean  &  82.5 & 79.3 & 85.7 & 82.7 & 81.1 & 78.2 \\
 &  & AR \cite{ar} &  80.1 & 77.1 & 83.2 & 80.6 & 79.9 & 75.0 \\
  &  & LSP \cite{lsp} &  75.9 & 73.0 & 78.7 & 77.7 & 77.0 & 73.0 \\
    &  & EM \cite{em} &  71.5 & 68.4 & 74.6 & 76.0 & 74.8 & 72.4 \\
&  & \textbf{TUE (Ours)}  &  \textbf{63.6 (18.9$\downarrow$)} & \textbf{59.9 (19.4$\downarrow$)} & \textbf{67.4 (18.3$\downarrow$)} & \textbf{65.7} (\textbf{17.0}$\downarrow$) & \textbf{62.6} (\textbf{18.5}$\downarrow$) & \textbf{62.2} (\textbf{16.0}$\downarrow$) \\
 \midrule
 \multirow{5}{*}{XMEM \cite{XMEM}} &  \multirow{5}{*}{DAVIS-17+YTVOS-19} & Clean & 84.5 & 81.4 & 87.6 &  84.2 & 83.8 & 78.1 \\
  & & AR \cite{ar} & 82.1 & 78.7 & 83.9 & 81.8 & 81.9 & 75.3 \\ 
 & & LSP \cite{lsp} & 81.7 & 78.6 & 84.7 & 82.0 & 80.9& 76.9 \\ 
  & & EM \cite{em} & 78.9 & 75.7 & 82.1 & 80.5 & 78.3 & 76.6 \\ 
 & & \textbf{TUE (Ours)} & \textbf{64.2 (20.3$\downarrow$)} & \textbf{60.5 (20.9$\downarrow$)} & \textbf{67.9 (19.7$\downarrow$)} & \textbf{61.3} (\textbf{22.9}$\downarrow$) & \textbf{58.4} (\textbf{25.4}$\downarrow$) & \textbf{58.1} (\textbf{20.0}$\downarrow$)  \\ 
\bottomrule
\end{tabular} 
}
 \vspace{-0.3cm}
\caption{Transfer to dense temporal matching task, i.e., Video Object Segmentation (VOS). We apply our TUE generator, trained with SiamFC on GOT-10k for the VOT task, to perform zero-shot TUEs generation on DAVIS-17 and YTVOS-19 training sets. The mask annotations are firstly converted to box annotations. Lower performance indicates stronger training data privacy protection.}
\label{task_transfer}
 \vspace{-0.5cm}
\end{table*}

\noindent \textbf{Transfer to State-of-the-art Trackers.} We train our TUE generator with the naive SiamFC tracker on GOT-10k \cite{got10k} and use the learned generator to produce TUE-perturbed GOT-10k (TUE-GOT10k). To evaluate the transferability, we train state-of-the-art trackers, including OSTrack-256 \cite{ostrack}, DropTrack-384 \cite{dropmae}, SeqTrack-256 \cite{chen2023seqtrack}, MixFormer-CvT \cite{MixFormer}, STARK-S50 \cite{stark}, AQATrack-256 \cite{aqatrack}, and HIPTrack \cite{hiptrack}, on TUE-GOT10k, following their original training settings.  
For other UE methods: (1) EM \cite{em} and TAP \cite{tap} are optimization-based approaches specifically optimized with SiamFC on GOT-10k; (2) LSP \cite{lsp} and AR \cite{ar} are class-wise UEs. To adapt them, we randomly sample a class-wise UE noise per video, creating perturbed videos. More details are provided in the Supplementary.  


Tab. \ref{overall_results} shows that our TUE outperforms other UE methods in privacy protection across various tracking architectures, including CNNs (SiamFC, Stark), ViTs (OSTrack, DropTrack), and Decoders (HIPTrack, AQATrack). 
HIPTrack exhibits less performance degradation since it uses DropTrack as the frozen base tracker, which limits its ability to learn shortcut features. 
Notably, our TUEs, trained with a naive SiamFC model in about seven hours on a single NVIDIA 4090 GPU, transfer effectively to SOTA trackers, eliminating the need for costly optimization with complex tracking models.

\begin{figure}[t]
\vspace{-0.2cm}
  \centering
  \includegraphics[width=0.84\linewidth]{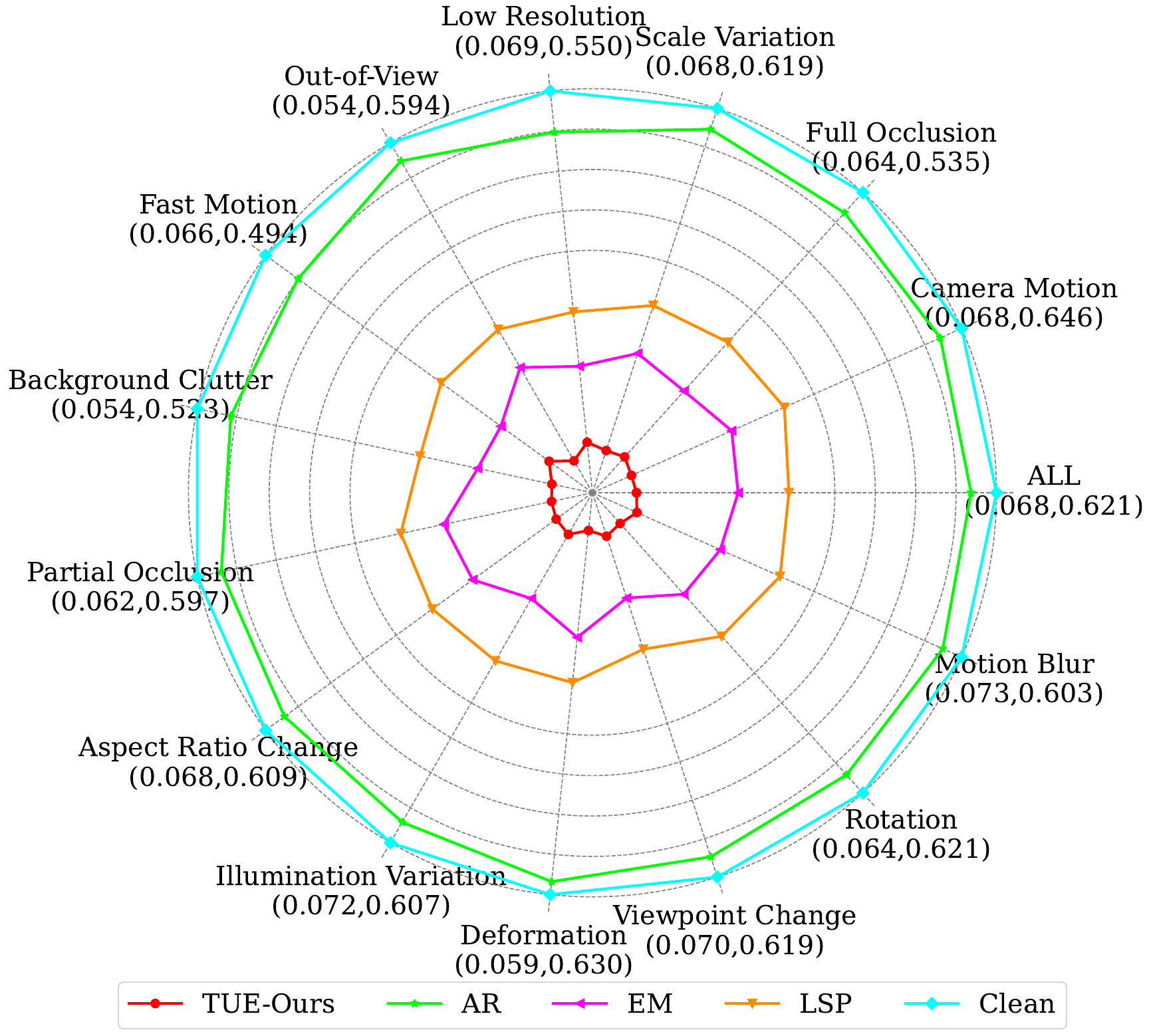}
  \vspace{-3.5mm}
   \caption{AUC scores of different attributes on LaSOT \cite{lasot}.  
   We use STARK-S50 \cite{stark} as the base tracker, which is trained with perturbed datasets (i.e., LaSOT + GOT-10k \cite{got10k}) generated by AR, EM, LSP, and our TUE. Best viewed in color.}
   \label{lasot_attr}
   \vspace{-5mm}
\end{figure}

\noindent \textbf{Transfer to Unseen Datasets.} To ensure scalability on large-scale video datasets, we perform zero-shot TUE generation on unseen datasets. Specifically, after training the TUE generator on GOT-10k w/ SiamFC, we use it to generate TUEs for LaSOT (TUE-LaSOT) without additional training. We then train STARK-S50 \cite{stark} on TUE-LaSOT for evaluation. 
As shown in Tab.~\ref{dataset_transfer}, the optimization-based EM struggles to generalize beyond GOT-10k, requiring video-wise UEs sampled from GOT-10k for LaSOT. In contrast, our TUE achieves significant performance drops across three datasets. Training STARK-S50 on both TUE-LaSOT and TUE-GOT10k further amplifies performance degradation. Overall, our method efficiently generates large-scale TUEs via model inference without retraining.

\noindent \textbf{Transfer to Video Object Segmentation (VOS).} VOS is a dense temporal matching task that segments targets in each frame based on initial mask annotations. We evaluate our TUEs in this context by first generating bounding boxes around mask annotations and then using a generator trained on VOT to create unlearnable VOS datasets, TUE-DAVIS17 and TUE-YTVOS19. These datasets are used to train existing VOS methods following standard protocols.
As shown in Tab.~\ref{task_transfer}, despite differences between VOT and VOS, our TUEs cause significant performance degradation across various VOS methods and datasets, highlighting their effectiveness in disrupting temporal matching and protecting video data privacy. {Further transferability to long-term point tracking is demonstrated in the supplementary.}

\noindent \textbf{Attribute Analysis.} Fig.~\ref{lasot_attr} presents the AUC scores for various attributes on LaSOT~\cite{lasot}. First, we apply AR, EM, LSP, and our TUE methods to the VOT dataset (i.e., LaSOT~\cite{lasot} + GOT-10k \cite{got10k}) to generate unlearnable datasets. We then use STARK-S50 \cite{stark} as the base tracker, training it on the unlearnable datasets produced by these data privacy protection algorithms. Our method shows largest performance drop across all attributes.


\begin{figure}[t]
\vspace{-0.1cm}
  \centering
  \includegraphics[width=1.0\linewidth]{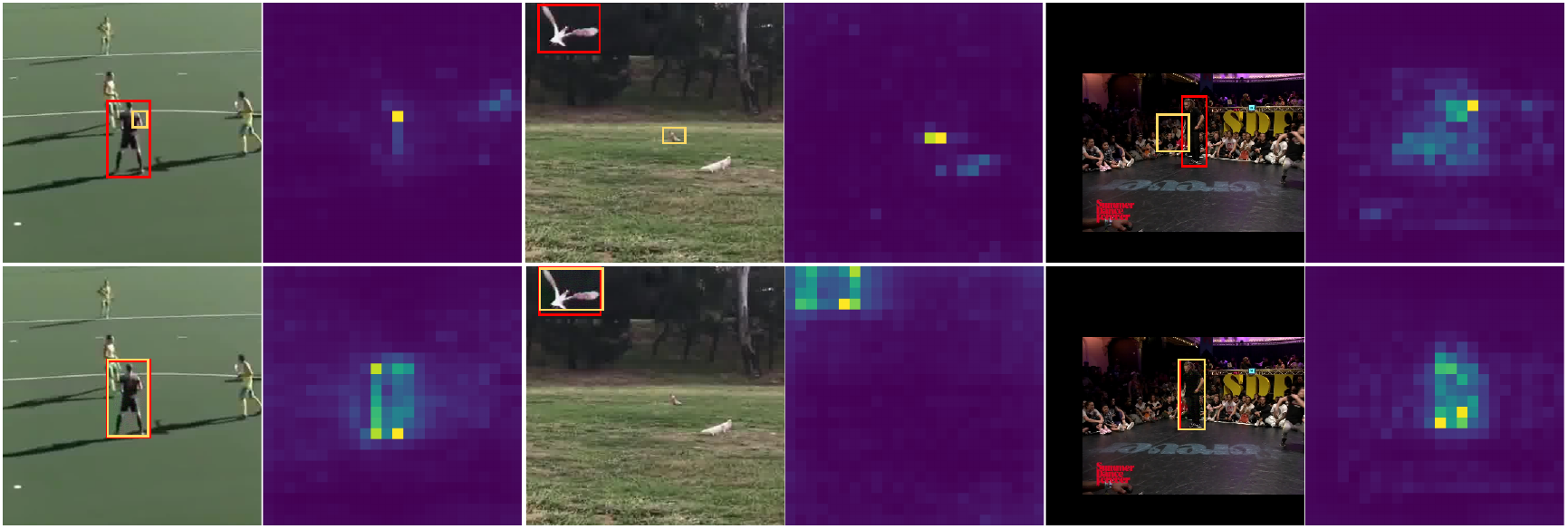}
  \vspace{-7mm}
   \caption{Template-to-search attention visualization from TUE-DropTrack on clean videos (top) and TUE-perturbed videos (bottom). Red and yellow rectangles are ground truth and predicted bounding boxes in search regions, respectively.}
   \label{vis_attn}
   \vspace{-5mm}
\end{figure}

\noindent \textbf{Visualization.} Fig.~\ref{vis_attn} visualizes the template-to-search attention weights from the last layer of ViT in TUE-DropTrack. TUE-DropTrack, trained with TUEs, heavily relies on them for temporal matching. For clean videos, TUE-DropTrack generates inaccurate attention maps due to overfitting on TUEs. 
{The supplementary provides additional visualizations demonstrating that the perturbations are imperceptible while maintaining high-quality perturbed frames, along with further visualizations of attention weights, perturbations, and TUEs.
}



\vspace{-2mm}
\section{Conclusion}
\vspace{-2mm}
This paper presented the first effort to address the privacy concerns about unauthorized data exploitation in VOT. We constructed a comprehensive benchmark to evaluate existing UE methods, revealing prior methods' limitations in efficiency, effectiveness, and generalizability. To overcome these issues, we introduced Temporal Unlearnable Examples (TUEs) with a lightweight generative framework, which conditions on target states to produce target-aware noise perturbations. Additionally, we designed a temporal contrastive loss to encourage trackers to rely on TUEs during training, further strengthening privacy protection.  
Extensive experiments show that TUEs transfer well across trackers, datasets, and temporal matching tasks. Ablation studies validated the effectiveness and efficiency of our generative framework. Qualitative results confirm that the generated perturbations are imperceptible, effectively protect VOT videos from unauthorized use. We expect our TUE to advance data privacy in the tracking community.



\vspace{-2mm}
\section*{Acknowledgments}
\vspace{-2mm}
This work was supported by a grant from the Research Grants Council of the Hong Kong Special Administrative Region, China (CityU 11211624). This work was supported by the National Natural Science Foundation of China under Project 62406090. This work was carried out at the Rapid-Rich Object Search (ROSE) Lab, School of Electrical \& Electronic Engineering, Nanyang Technological University (NTU), Singapore. The research is supported by the National Research Foundation, Singapore and Infocomm Media Development Authority under its Trust Tech Funding Initiative. Any opinions, findings and conclusions or recommendations expressed in this material are those of the author(s) and do not reflect the views of National Research Foundation, Singapore and Infocomm Media Development Authority.



{
    \small
    \bibliographystyle{ieeenat_fullname}
    \bibliography{main}

\begin{thebibliography}{96}
\providecommand{\natexlab}[1]{#1}
\providecommand{\url}[1]{\texttt{#1}}
\expandafter\ifx\csname urlstyle\endcsname\relax
  \providecommand{\doi}[1]{doi: #1}\else
  \providecommand{\doi}{doi: \begingroup \urlstyle{rm}\Url}\fi

\bibitem[Abadi et~al.(2016)Abadi, Chu, Goodfellow, McMahan, Mironov, Talwar, and Zhang]{abadi2016deep}
Martin Abadi, Andy Chu, Ian Goodfellow, H~Brendan McMahan, Ilya Mironov, Kunal Talwar, and Li Zhang.
\newblock Deep learning with differential privacy.
\newblock In \emph{Proceedings of the 2016 ACM SIGSAC conference on computer and communications security}, pages 308--318, 2016.

\bibitem[Bai et~al.(2024)Bai, Zhao, Gong, and Wei]{artrack_v2}
Yifan Bai, Zeyang Zhao, Yihong Gong, and Xing Wei.
\newblock Artrackv2: Prompting autoregressive tracker where to look and how to describe.
\newblock In \emph{Proceedings of the IEEE/CVF Conference on Computer Vision and Pattern Recognition (CVPR)}, 2024.

\bibitem[Bertinetto et~al.(2016{\natexlab{a}})Bertinetto, Valmadre, Golodetz, Miksik, and Torr]{Staple}
L. Bertinetto, J. Valmadre, S. Golodetz, O. Miksik, and P. Torr.
\newblock Staple: Complementary learners for real-time tracking.
\newblock In \emph{CVPR}, pages 1401--1409, 2016{\natexlab{a}}.

\bibitem[Bertinetto et~al.(2016{\natexlab{b}})Bertinetto, Valmadre, Henriques, Vedaldi, and Vedaldi]{SiamFC}
L. Bertinetto, J. Valmadre, J.F. Henriques, A. Vedaldi, and P.H.S. Vedaldi.
\newblock Fully-convolutional siamese networks for object tracking.
\newblock In \emph{ECCV Workshop}, pages 850--865, 2016{\natexlab{b}}.

\bibitem[Cai et~al.(2024)Cai, Liu, and Wang]{hiptrack}
Wenrui Cai, Qingjie Liu, and Yunhong Wang.
\newblock Hiptrack: Visual tracking with historical prompts.
\newblock In \emph{Proceedings of the IEEE/CVF Conference on Computer Vision and Pattern Recognition}, 2024.

\bibitem[Chen et~al.(2020{\natexlab{a}})Chen, Kornblith, and Norouzi]{simclr}
T. Chen, S. Kornblith, and M. Norouzi.
\newblock A simple framework for contrastive learning of visual representations.
\newblock In \emph{International Conference on Machine Learning}, 2020{\natexlab{a}}.

\bibitem[Chen et~al.(2020{\natexlab{b}})Chen, Yan, Zheng, Jiang, Xia, Zhao, and Ji]{chen2020one}
Xuesong Chen, Xiyu Yan, Feng Zheng, Yong Jiang, Shu-Tao Xia, Yong Zhao, and Rongrong Ji.
\newblock One-shot adversarial attacks on visual tracking with dual attention.
\newblock In \emph{Proceedings of the IEEE/CVF conference on computer vision and pattern recognition}, pages 10176--10185, 2020{\natexlab{b}}.

\bibitem[Chen et~al.(2021)Chen, Yan, Zhu, Wang, Yang, and Lu]{transt}
X. Chen, B. Yan, J. Zhu, D. Wang, X. Yang, and H. Lu.
\newblock Transformer tracking.
\newblock In \emph{CVPR}, pages 8126--8135, 2021.

\bibitem[Chen et~al.(2023)Chen, Peng, Wang, Lu, and Hu]{chen2023seqtrack}
Xin Chen, Houwen Peng, Dong Wang, Huchuan Lu, and Han Hu.
\newblock Seqtrack: Sequence to sequence learning for visual object tracking.
\newblock In \emph{Proceedings of the IEEE/CVF conference on computer vision and pattern recognition}, pages 14572--14581, 2023.

\bibitem[Cheng and Schwing(2022)]{XMEM}
H.~K. Cheng and A~G. Schwing.
\newblock Xmem: Long-term video object segmentation with an atkinson-shiffrin memory model.
\newblock In \emph{ECCV}, 2022.

\bibitem[Cheng et~al.(2021)Cheng, Tai, and Tang]{STCN}
H.~K. Cheng, Y.~W. Tai, and C.~K. Tang.
\newblock Rethinking space-time networks with improved memory coverage for efficient video object segmentation.
\newblock In \emph{NeurIPS}, pages 11781--11794, 2021.

\bibitem[Cui et~al.(2022)Cui, Jiang, Wang, and Wu]{MixFormer}
Yutao Cui, Cheng Jiang, Limin Wang, and Gangshan Wu.
\newblock Mixformer: End-to-end tracking with iterative mixed attention.
\newblock In \emph{CVPR}, 2022.

\bibitem[Danelljan et~al.(2015{\natexlab{a}})Danelljan, Hager, Khan, and Felsberg]{DeepSRDCF}
M. Danelljan, G. Hager, F.~S. Khan, and M. Felsberg.
\newblock Convolutional features for correlation filter based visual tracking.
\newblock In \emph{ICCVW}, pages 58--66, 2015{\natexlab{a}}.

\bibitem[Danelljan et~al.(2015{\natexlab{b}})Danelljan, Hager, Khan, and Felsberg]{SRDCF}
M. Danelljan, G. Hager, F.~S. Khan, and M. Felsberg.
\newblock Learning spatially regularized correlation filters for visual tracking.
\newblock In \emph{ICCV}, pages 4310--4318, 2015{\natexlab{b}}.

\bibitem[Danelljan et~al.(2016)Danelljan, Robinson, Khan, and Felsberg]{CCOT}
M. Danelljan, A. Robinson, F.~S. Khan, and M. Felsberg.
\newblock Beyond correlation filters: learning continuous convolution operators for visual tracking.
\newblock In \emph{ECCV}, pages 472--488, 2016.

\bibitem[Danelljan et~al.(2017{\natexlab{a}})Danelljan, Bhat, Khan, and Felsberg]{ECO}
M. Danelljan, G. Bhat, F.~S. Khan, and M. Felsberg.
\newblock Eco: Efficient convolution operators for tracking.
\newblock In \emph{CVPR}, pages 21--26, 2017{\natexlab{a}}.

\bibitem[Danelljan et~al.(2017{\natexlab{b}})Danelljan, Häger, Khan, and Felsberg]{DSST}
M. Danelljan, G. Häger, F.~S. Khan, and M. Felsberg.
\newblock Discriminative scale space tracking.
\newblock \emph{IEEE Transactions on Pattern Analysis and Machine Intelligence}, 39\penalty0 (8):\penalty0 1561--1575, 2017{\natexlab{b}}.

\bibitem[Dosovitskiy et~al.(2021)Dosovitskiy, Beyer, and Kolesnikov]{ViT}
A. Dosovitskiy, L. Beyer, and A. Kolesnikov.
\newblock An image is worth 16x16 words: Transformers for image recognition at scale.
\newblock In \emph{ICLR}, 2021.

\bibitem[Fan et~al.(2019)Fan, Lin, and Yang]{lasot}
H. Fan, L. Lin, and F. Yang.
\newblock Lasot: A high-quality benchmark for large-scale single object tracking.
\newblock In \emph{CVPR}, pages 5374--5383, 2019.

\bibitem[Fan et~al.(2021)Fan, Bai, Lin, Yang, Chu, Deng, Yu, Huang, Liu, and Xu]{lasotext}
H. Fan, H. Bai, L. Lin, F. Yang, P. Chu, G. Deng, S. Yu, M. Huang, J. Liu, and Y. Xu.
\newblock Lasot: A high-quality large-scale single object tracking benchmark.
\newblock In \emph{IJCV}, 2021.

\bibitem[Feng et~al.(2019)Feng, Cai, and Zhou]{dc}
Ji Feng, Qi-Zhi Cai, and Zhi-Hua Zhou.
\newblock Learning to confuse: generating training time adversarial data with auto-encoder.
\newblock \emph{NeurIPS}, 32, 2019.

\bibitem[Fowl et~al.(2021)Fowl, Goldblum, Chiang, Geiping, Czaja, and Goldstein]{tap}
Liam Fowl, Micah Goldblum, Ping-yeh Chiang, Jonas Geiping, Wojciech Czaja, and Tom Goldstein.
\newblock Adversarial examples make strong poisons.
\newblock \emph{NeurIPS}, 34:\penalty0 30339--30351, 2021.

\bibitem[Fu et~al.(2022)Fu, He, Liu, Shen, and Tao]{rem}
Shaopeng Fu, Fengxiang He, Yang Liu, Li Shen, and Dacheng Tao.
\newblock Robust unlearnable examples: Protecting data privacy against adversarial learning.
\newblock In \emph{ICLR}, 2022.

\bibitem[Galoogahi et~al.(2017)Galoogahi, Fagg, and Lucey]{BACF}
H. Galoogahi, A. Fagg, and S. Lucey.
\newblock Learning background-aware correlation filters for visual tracking.
\newblock In \emph{ICCV}, 2017.

\bibitem[Gao et~al.(2023)Gao, Zhou, and Zhang]{gao2023generalized}
Shenyuan Gao, Chunluan Zhou, and Jun Zhang.
\newblock Generalized relation modeling for transformer tracking.
\newblock In \emph{Proceedings of the IEEE/CVF Conference on Computer Vision and Pattern Recognition}, pages 18686--18695, 2023.

\bibitem[Guo et~al.(2020)Guo, Xie, Juefei-Xu, Ma, Li, Xue, Feng, and Liu]{spark}
Qing Guo, Xiaofei Xie, Felix Juefei-Xu, Lei Ma, Zhongguo Li, Wanli Xue, Wei Feng, and Yang Liu.
\newblock Spark: Spatial-aware online incremental attack against visual tracking.
\newblock In \emph{European conference on computer vision}, pages 202--219. Springer, 2020.

\bibitem[He et~al.(2016)He, Zhang, Ren, and Sun]{resnet}
Kaiming He, Xiangyu Zhang, Shaoqing Ren, and Jian Sun.
\newblock Deep residual learning for image recognition.
\newblock In \emph{CVPR}, pages 770--778, 2016.

\bibitem[Henriques et~al.(2015)Henriques, Caseiro, Martins, and Batista]{KCF}
J.~F. Henriques, R. Caseiro, P. Martins, and J. Batista.
\newblock High-speed tracking with kernelized correlation filters.
\newblock \emph{IEEE Transactions on Pattern Analysis and Machine Intelligence}, 37\penalty0 (3):\penalty0 583--596, 2015.

\bibitem[Hong et~al.(2024)Hong, Yan, Zhang, Li, Zhou, Guo, Jiang, Chen, Li, and Chen]{hong2024onetracker}
Lingyi Hong, Shilin Yan, Renrui Zhang, Wanyun Li, Xinyu Zhou, Pinxue Guo, Kaixun Jiang, Yiting Chen, Jinglun Li, and Zhaoyu Chen.
\newblock Onetracker: Unifying visual object tracking with foundation models and efficient tuning.
\newblock In \emph{Proceedings of the IEEE/CVF Conference on Computer Vision and Pattern Recognition}, 2024.

\bibitem[Huang et~al.(2024)Huang, Yu, Chen, Pan, Wang, and Wang]{huang2024badtrack}
Bin Huang, Jiaqian Yu, Yiwei Chen, Siyang Pan, Qiang Wang, and Zhi Wang.
\newblock Badtrack: a poison-only backdoor attack on visual object tracking.
\newblock \emph{Advances in Neural Information Processing Systems}, 36, 2024.

\bibitem[Huang et~al.(2021)Huang, Ma, Erfani, Bailey, and Wang]{em}
Hanxun Huang, Xingjun Ma, Sarah~Monazam Erfani, James Bailey, and Yisen Wang.
\newblock Unlearnable examples: Making personal data unexploitable.
\newblock In \emph{ICLR}, 2021.

\bibitem[Huang et~al.(2019)Huang, Zhao, and Huang]{got10k}
L. Huang, X. Zhao, and K. Huang.
\newblock Got-10k: A large high-diversity benchmark for generic object tracking in the wild.
\newblock \emph{IEEE Transactions on Pattern Analysis and Machine Intelligence}, 2019.

\bibitem[Jia et~al.(2020)Jia, Ma, Song, and Yang]{jia2020robust}
Shuai Jia, Chao Ma, Yibing Song, and Xiaokang Yang.
\newblock Robust tracking against adversarial attacks.
\newblock In \emph{Computer Vision--ECCV 2020: 16th European Conference, Glasgow, UK, August 23--28, 2020, Proceedings, Part XIX 16}, pages 69--84. Springer, 2020.

\bibitem[Kingma and Ba(2014)]{Adam}
Diederik~P. Kingma and Jimmy~Lei Ba.
\newblock Adam: A method for stochastic optimization.
\newblock In \emph{arXiv:1412.6980}, 2014.

\bibitem[Krizhevsky et~al.(2017)Krizhevsky, Ilya, and Hinton]{alex}
A. Krizhevsky, S. Ilya, and G. Hinton.
\newblock Imagenet classification with deep convolutional neural networks.
\newblock In \emph{Communications of the ACM}, pages 84--90, 2017.

\bibitem[Li et~al.(2018)Li, Wu, Zhu, and Yan]{siamrpn}
B. Li, W. Wu, Z. Zhu, and J. Yan.
\newblock High performance visual tracking with siamese region proposal network.
\newblock In \emph{Proceedings of the CVPR}, pages 8971--8980, 2018.

\bibitem[Li et~al.(2019)Li, Wu, Wang, Zhang, Xing, and Yan]{SiamRPN_plus}
B. Li, W. Wu, Q. Wang, F. Zhang, J. Xing, and J. Yan.
\newblock Siamrpn++: Evolution of siamese visual tracking with very deep networks.
\newblock In \emph{CVPR}, 2019.

\bibitem[Li et~al.(2023)Li, Chen, and Xing]{li2023memory}
Jiahao Li, Yiqiang Chen, and Yunbing Xing.
\newblock Memory mechanism for unsupervised anomaly detection.
\newblock In \emph{Uncertainty in Artificial Intelligence}, pages 1219--1229. PMLR, 2023.

\bibitem[Li et~al.(2024)Li, Chen, Xing, Gu, and Lan]{li2024cascade}
Jiahao Li, Yiqiang Chen, Yunbing Xing, Yang Gu, and Xiangyuan Lan.
\newblock Cascade memory for unsupervised anomaly detection.
\newblock In \emph{ECAI}, pages 2854--2861. IOS Press, 2024.

\bibitem[Li et~al.(2025{\natexlab{a}})Li, Chen, Xing, Gu, and Lan]{li2025contrast}
Jiahao Li, Yiqiang Chen, Yunbing Xing, Yang Gu, and Xiangyuan Lan.
\newblock Contrast memory for unsupervised anomaly detection.
\newblock In \emph{IEEE International Conference on Acoustics, Speech and Signal Processing}, pages 1--5. IEEE, 2025{\natexlab{a}}.

\bibitem[Li et~al.(2025{\natexlab{b}})Li, Chen, Xing, Gu, and Lan]{li2025hyman}
Jiahao Li, Yiqiang Chen, Yunbing Xing, Yang Gu, and Xiangyuan Lan.
\newblock Hyman: Hybrid memory and attention network for unsupervised anomaly detection.
\newblock In \emph{IEEE International Conference on Acoustics, Speech and Signal Processing}, pages 1--5. IEEE, 2025{\natexlab{b}}.

\bibitem[Li et~al.(2025{\natexlab{c}})Li, Chen, Xing, Gu, and Lan]{li2025kbs}
Jiahao Li, Yiqiang Chen, Yunbing Xing, Yang Gu, and Xiangyuan Lan.
\newblock K-space bispectrum steganography for robust unlearnable data.
\newblock In \emph{Proceedings of the 33rd ACM International Conference on Multimedia}. Association for Computing Machinery, 2025{\natexlab{c}}.

\bibitem[Li et~al.(2025{\natexlab{d}})Li, Chen, Xing, Gu, and Lan]{li2025surveyunlearnabledata}
Jiahao Li, Yiqiang Chen, Yunbing Xing, Yang Gu, and Xiangyuan Lan.
\newblock A survey on unlearnable data, 2025{\natexlab{d}}.

\bibitem[Li et~al.(2022)Li, Zhong, Ma, Jiang, and Xia]{li2022few}
Yiming Li, Haoxiang Zhong, Xingjun Ma, Yong Jiang, and Shu-Tao Xia.
\newblock Few-shot backdoor attacks on visual object tracking.
\newblock \emph{arXiv preprint arXiv:2201.13178}, 2022.

\bibitem[Liang et~al.(2020)Liang, Wei, Yao, and Cao]{liang2020efficient}
Siyuan Liang, Xingxing Wei, Siyuan Yao, and Xiaochun Cao.
\newblock Efficient adversarial attacks for visual object tracking.
\newblock In \emph{ECCV}, pages 34--50, 2020.

\bibitem[Liang et~al.(2021)Liang, Wu, Liu, Yan, and Wang]{robust_cf_journal}
Y. Liang, Q. Wu, Y. Liu, Y. Yan, and H. Wang.
\newblock Deep correlation filter tracking with shepherded instance-aware proposals.
\newblock In \emph{IEEE Transactions on Intelligent Transportation Systems}, 2021.

\bibitem[Lin et~al.(2024{\natexlab{a}})Lin, Tang, Wang, Liu, Ju, Wang, and Yu]{DBLP:journals/patterns/LinTWLJWY24}
Xun Lin, Wenzhong Tang, Haoran Wang, Yizhong Liu, Yakun Ju, Shuai Wang, and Zitong Yu.
\newblock Exposing image splicing traces in scientific publications via uncertainty-guided refinement.
\newblock \emph{Patterns}, 5\penalty0 (9):\penalty0 101038, 2024{\natexlab{a}}.

\bibitem[Lin et~al.(2024{\natexlab{b}})Lin, Yu, Xia, Jiang, Wang, Yu, Liu, Fu, Wang, Tang, et~al.]{lin2024safeguarding}
Xun Lin, Yi Yu, Song Xia, Jue Jiang, Haoran Wang, Zitong Yu, Yizhong Liu, Ying Fu, Shuai Wang, Wenzhong Tang, et~al.
\newblock Safeguarding medical image segmentation datasets against unauthorized training via contour-and texture-aware perturbations.
\newblock \emph{arXiv preprint arXiv:2403.14250}, 2024{\natexlab{b}}.

\bibitem[Lin et~al.(2024{\natexlab{c}})Lin, Yu, Yu, Meng, Zhou, Liu, Liu, Wang, Tang, Lei, et~al.]{lin2024hidemia}
Xun Lin, Yi Yu, Zitong Yu, Ruohan Meng, Jiale Zhou, Ajian Liu, Yizhong Liu, Shuai Wang, Wenzhong Tang, Zhen Lei, et~al.
\newblock Hidemia: Hidden wavelet mining for privacy-enhancing medical image analysis.
\newblock In \emph{ACM MM}, pages 8110--8119, 2024{\natexlab{c}}.

\bibitem[Liu et~al.(2022)Liu, Liang, Wu, Zhang, and Wang]{liu2022new}
Yi Liu, Yanjie Liang, Qiangqiang Wu, Liming Zhang, and Hanzi Wang.
\newblock A new framework for multiple deep correlation filters based object tracking.
\newblock In \emph{IEEE International Conference on Acoustics, Speech and Signal Processing}, pages 1670--1674, 2022.

\bibitem[Liu et~al.(2023)Liu, Zhao, and Larson]{iss}
Zhuoran Liu, Zhengyu Zhao, and Martha Larson.
\newblock Image shortcut squeezing: Countering perturbative availability poisons with compression.
\newblock \emph{International Conference on Machine Learning}, 2023.

\bibitem[Lukezic and Vojir(2017)]{CSRDCF}
A. Lukezic and T. Vojir.
\newblock Discriminative correlation filter woth channel and spatial reliability.
\newblock In \emph{CVPR}, 2017.

\bibitem[Madry et~al.(2018)Madry, Makelov, Schmidt, Tsipras, and Vladu]{pgd}
Aleksander Madry, Aleksandar Makelov, Ludwig Schmidt, Dimitris Tsipras, and Adrian Vladu.
\newblock Towards deep learning models resistant to adversarial attacks.
\newblock In \emph{ICLR}, 2018.

\bibitem[Meng et~al.(2024)Meng, Yi, Yu, Yang, Shen, and Kot]{meng2024semantic}
Ruohan Meng, Chenyu Yi, Yi Yu, Siyuan Yang, Bingquan Shen, and Alex~C Kot.
\newblock Semantic deep hiding for robust unlearnable examples.
\newblock \emph{IEEE Transactions on Information Forensics and Security}, 2024.

\bibitem[Muller et~al.(2018)Muller, Bibi, and S]{trackingnet}
M. Muller, A. Bibi, and Giancola S.
\newblock Trackingnet: A large-scale dataset and benchmark for object tracking in the wild.
\newblock In \emph{ECCV}, pages 300--317, 2018.

\bibitem[Nakka and Salzmann(2020)]{nakka2020temporally}
Krishna~Kanth Nakka and Mathieu Salzmann.
\newblock Temporally-transferable perturbations: Efficient, one-shot adversarial attacks for online visual object trackers.
\newblock \emph{arXiv preprint arXiv:2012.15183}, 2020.

\bibitem[Peebles and Xie(2023)]{dit}
William Peebles and Saining Xie.
\newblock Scalable diffusion models with transformers.
\newblock In \emph{Proceedings of the IEEE/CVF International Conference on Computer Vision}, pages 4195--4205, 2023.

\bibitem[Phan et~al.(2016)Phan, Wang, Wu, and Dou]{phan2016differential}
NhatHai Phan, Yue Wang, Xintao Wu, and Dejing Dou.
\newblock Differential privacy preservation for deep auto-encoders: an application of human behavior prediction.
\newblock In \emph{Proceedings of the AAAI Conference on Artificial Intelligence}, 2016.

\bibitem[Pont-Tuset et~al.(2017)Pont-Tuset, Perazzi, Caelles, Arbeláez, Sorkine-Hornung, and Van~Gool]{davis17}
J. Pont-Tuset, F. Perazzi, S. Caelles, P. Arbeláez, A. Sorkine-Hornung, and L. Van~Gool.
\newblock The 2017 davis challenge on video object segmentation.
\newblock In \emph{arXiv:1704.00675}, 2017.

\bibitem[Sandoval-Segura et~al.(2022)Sandoval-Segura, Singla, Geiping, Goldblum, Goldstein, and Jacobs]{ar}
Pedro Sandoval-Segura, Vasu Singla, Jonas Geiping, Micah Goldblum, Tom Goldstein, and David Jacobs.
\newblock Autoregressive perturbations for data poisoning.
\newblock \emph{NeurIPS}, 35:\penalty0 27374--27386, 2022.

\bibitem[Shan et~al.(2020)Shan, Wenger, Zhang, Li, Zheng, and Zhao]{shan2020fawkes}
Shawn Shan, Emily Wenger, Jiayun Zhang, Huiying Li, Haitao Zheng, and Ben~Y Zhao.
\newblock Fawkes: Protecting privacy against unauthorized deep learning models.
\newblock In \emph{29th USENIX security symposium (USENIX Security 20)}, pages 1589--1604, 2020.

\bibitem[Shokri and Shmatikov(2015)]{shokri2015privacy}
Reza Shokri and Vitaly Shmatikov.
\newblock Privacy-preserving deep learning.
\newblock In \emph{Proceedings of the 22nd ACM SIGSAC conference on computer and communications security}, pages 1310--1321, 2015.

\bibitem[Shokri et~al.(2017)Shokri, Stronati, Song, and Shmatikov]{shokri2017membership}
Reza Shokri, Marco Stronati, Congzheng Song, and Vitaly Shmatikov.
\newblock Membership inference attacks against machine learning models.
\newblock In \emph{2017 IEEE symposium on security and privacy (SP)}, pages 3--18. IEEE, 2017.

\bibitem[Sun et~al.(2024)Sun, Zhang, Zhang, Ma, and Jiang]{sun2024unseg}
Ye Sun, Hao Zhang, Tiehua Zhang, Xingjun Ma, and Yu-Gang Jiang.
\newblock Unseg: One universal unlearnable example generator is enough against all image segmentation.
\newblock \emph{arXiv preprint arXiv:2410.09909}, 2024.

\bibitem[Sun et~al.(2021)Sun, Chen, Chao, Ruan, and Mukherjee]{9142255}
Zhihong Sun, Jun Chen, Liang Chao, Weijian Ruan, and Mithun Mukherjee.
\newblock A survey of multiple pedestrian tracking based on tracking-by-detection framework.
\newblock \emph{IEEE Transactions on Circuits and Systems for Video Technology}, 31\penalty0 (5):\penalty0 1819--1833, 2021.

\bibitem[Tao et~al.(2016)Tao, Gavves, and Smeulders]{SINT}
R. Tao, E. Gavves, and A.~W.M. Smeulders.
\newblock Siamese instance search for tracking.
\newblock In \emph{CVPR}, pages 1420--1429, 2016.

\bibitem[Valmadre et~al.(2017)Valmadre, Bertinetto, Henriques, Vedaldi, and Torr]{CFNet}
J. Valmadre, L. Bertinetto, J. Henriques, A. Vedaldi, and P.~H.~S. Torr.
\newblock End-to-end representation learning for correlation filter based tracking.
\newblock In \emph{CVPR}, pages 5000--5008, 2017.

\bibitem[Wang et~al.(2017)Wang, Gao, and Xing]{DCFNet}
Q. Wang, J. Gao, and J. Xing.
\newblock Dcfnet: Discriminant correlation filters network for visual tracking.
\newblock In \emph{arXiv:1704.04057}, 2017.

\bibitem[Wang et~al.(2021)Wang, Shu, Zhang, Jiang, Wang, Tian, and Wu]{tnl2k}
X. Wang, X. Shu, Z. Zhang, B. Jiang, Y. Wang, Y. Tian, and F. Wu.
\newblock Towards more flexible and accurate object tracking with natural language: Algorithms and benchmark.
\newblock In \emph{CVPR}, 2021.

\bibitem[Wang et~al.(2024)Wang, Li, Liu, Zhang, Hu, Zhang, Zhou, and Jin]{wang2024unlearnable}
Xianlong Wang, Minghui Li, Wei Liu, Hangtao Zhang, Shengshan Hu, Yechao Zhang, Ziqi Zhou, and Hai Jin.
\newblock Unlearnable 3d point clouds: Class-wise transformation is all you need.
\newblock \emph{arXiv preprint arXiv:2410.03644}, 2024.

\bibitem[Wei et~al.(2023)Wei, Bai, Zheng, Shi, and Gong]{artrack}
Xing Wei, Yifan Bai, Yongchao Zheng, Dahu Shi, and Yihong Gong.
\newblock Autoregressive visual tracking.
\newblock In \emph{Proceedings of the IEEE/CVF Conference on Computer Vision and Pattern Recognition (CVPR)}, 2023.

\bibitem[Wu and Chan(2021)]{wu2021meta}
Qiangqiang Wu and Antoni~B Chan.
\newblock Meta-graph adaptation for visual object tracking.
\newblock In \emph{2021 IEEE International Conference on Multimedia and Expo (ICME)}, pages 1--6, 2021.

\bibitem[Wu et~al.(2018)Wu, Yan, Liang, Liu, and Wang]{dsnet}
Q. Wu, Y. Yan, Y. Liang, Y. Liu, and H. Wang.
\newblock Dsnet: Deep and shallow feature learning for efficient visual tracking.
\newblock In \emph{ACCV}, pages 119--134, 2018.

\bibitem[Wu et~al.(2019{\natexlab{a}})Wu, Yan, Liang, Liu, and Wang]{wu2019dsnet}
Qiangqiang Wu, Yan Yan, Yanjie Liang, Yi Liu, and Hanzi Wang.
\newblock Dsnet: Deep and shallow feature learning for efficient visual tracking.
\newblock In \emph{Asian Conference on Computer Vision}, pages 119--134, 2019{\natexlab{a}}.

\bibitem[Wu et~al.(2023{\natexlab{a}})Wu, Yang, Liu, Wu, Shan, and Chan]{dropmae}
Qiangqiang Wu, Tianyu Yang, Ziquan Liu, Baoyuan Wu, Ying Shan, and Antoni~B. Chan.
\newblock Dropmae: Masked autoencoders with spatial-attention dropout for tracking tasks.
\newblock In \emph{Proceedings of the IEEE/CVF Conference on Computer Vision and Pattern Recognition (CVPR)}, pages 14561--14571, 2023{\natexlab{a}}.

\bibitem[Wu et~al.(2023{\natexlab{b}})Wu, Yang, Wu, and Chan]{wu2023scalable}
Qiangqiang Wu, Tianyu Yang, Wei Wu, and Antoni~B Chan.
\newblock Scalable video object segmentation with simplified framework.
\newblock In \emph{Proceedings of the IEEE/CVF international conference on computer vision}, pages 13879--13889, 2023{\natexlab{b}}.

\bibitem[Wu et~al.(2023{\natexlab{c}})Wu, Chen, Xie, and Huang]{ops}
Shutong Wu, Sizhe Chen, Cihang Xie, and Xiaolin Huang.
\newblock One-pixel shortcut: On the learning preference of deep neural networks.
\newblock In \emph{ICLR}, 2023{\natexlab{c}}.

\bibitem[Wu et~al.(2019{\natexlab{b}})Wu, Wang, Zhou, and Jian]{sta}
Xugang Wu, Xiaoping Wang, Xu Zhou, and Songlei Jian.
\newblock Sta: Adversarial attacks on siamese trackers.
\newblock \emph{arXiv preprint arXiv:1909.03413}, 2019{\natexlab{b}}.

\bibitem[Wu et~al.(2015)Wu, Lim, and Yang]{OTB100}
Y. Wu, J. Lim, and M.-H. Yang.
\newblock Object tracking benchmark.
\newblock \emph{IEEE Transactions on Pattern Analysis and Machine Intelligence}, 37\penalty0 (9):\penalty0 1834--1848, 2015.

\bibitem[Xie et~al.(2024{\natexlab{a}})Xie, Wang, and Ma]{diffusiontrack}
Fei Xie, Zhongdao Wang, and Chao Ma.
\newblock Diffusiontrack: Point set diffusion model for visual object tracking.
\newblock In \emph{Proceedings of the IEEE/CVF Conference on Computer Vision and Pattern Recognition (CVPR)}, 2024{\natexlab{a}}.

\bibitem[Xie et~al.(2024{\natexlab{b}})Xie, Zhong, Mo, Zhang, Shi, Song, and Ji]{aqatrack}
Jinxia Xie, Bineng Zhong, Zhiyi Mo, Shengping Zhang, Liangtao Shi, Shuxiang Song, and Rongrong Ji.
\newblock Autoregressive queries for adaptive tracking with spatio-temporal transformers.
\newblock In \emph{Proceedings of the IEEE/CVF Conference on Computer Vision and Pattern Recognition}, pages 19300--19309, 2024{\natexlab{b}}.

\bibitem[Xu et~al.(2018)Xu, Yang, Fan, Yue, Liang, Yang, and Huang]{youtubevos}
N. Xu, L. Yang, Y. Fan, D. Yue, Y. Liang, J. Yang, and T. Huang.
\newblock Youtube-vos: A large-scale video object segmentation benchmark.
\newblock In \emph{arXiv:1809.03327}, 2018.

\bibitem[Yan et~al.(2020{\natexlab{a}})Yan, Wang, Lu, and Yang]{yan2020cooling}
Bin Yan, Dong Wang, Huchuan Lu, and Xiaoyun Yang.
\newblock Cooling-shrinking attack: Blinding the tracker with imperceptible noises.
\newblock In \emph{Proceedings of the IEEE/CVF conference on computer vision and pattern recognition}, pages 990--999, 2020{\natexlab{a}}.

\bibitem[Yan et~al.(2021)Yan, Peng, Fu, Wang, and Lu]{stark}
B. Yan, H. Peng, J. Fu, D. Wang, and H. Lu.
\newblock Learning spatio-temporal transformer for visual tracking.
\newblock In \emph{ICCV}, pages 10448--10457, 2021.

\bibitem[Yan et~al.(2020{\natexlab{b}})Yan, Chen, Jiang, Xia, Zhao, and Zheng]{yan2020hijacking}
Xiyu Yan, Xuesong Chen, Yong Jiang, Shu-Tao Xia, Yong Zhao, and Feng Zheng.
\newblock Hijacking tracker: A powerful adversarial attack on visual tracking.
\newblock In \emph{ICASSP}, pages 2897--2901, 2020{\natexlab{b}}.

\bibitem[Yang and Chan(2018)]{memtrack}
T. Yang and A.~B. Chan.
\newblock Learning dynamic memory networks for object tracking.
\newblock In \emph{ECCV}, pages 152--167, 2018.

\bibitem[Yang et~al.(2020)Yang, Xu, and Hu]{roam}
T. Yang, P. Xu, and R. Hu.
\newblock Roam: Recurrently optimizing tracking model.
\newblock In \emph{CVPR}, pages 6718--6727, 2020.

\bibitem[Ye et~al.(2022)Ye, Chang, Ma, and Shan]{ostrack}
B. Ye, H. Chang, B. Ma, and S. Shan.
\newblock Joint feature learning and relation modeling for tracking: A one-stream framework.
\newblock In \emph{ECCV}, pages 341--357, 2022.

\bibitem[Yu et~al.(2022)Yu, Zhang, Chen, Yin, and Liu]{lsp}
Da Yu, Huishuai Zhang, Wei Chen, Jian Yin, and Tie-Yan Liu.
\newblock Availability attacks create shortcuts.
\newblock In \emph{Proceedings of the 28th ACM SIGKDD Conference on Knowledge Discovery and Data Mining}, pages 2367--2376, 2022.

\bibitem[Yu et~al.(2024{\natexlab{a}})Yu, Wang, Xia, Yang, Lu, Tan, and Kot]{yu2024purify}
Yi Yu, Yufei Wang, Song Xia, Wenhan Yang, Shijian Lu, Yap-Peng Tan, and Alex~C Kot.
\newblock Purify unlearnable examples via rate-constrained variational autoencoders.
\newblock In \emph{International Conference on Machine Learning, {ICML} 2024}, 2024{\natexlab{a}}.

\bibitem[Yu et~al.(2024{\natexlab{b}})Yu, Zheng, Yang, Yang, Liu, Lu, Tan, Lam, and Kot]{yu2024unlearnable}
Yi Yu, Qichen Zheng, Siyuan Yang, Wenhan Yang, Jun Liu, Shijian Lu, Yap-Peng Tan, Kwok-Yan Lam, and Alex Kot.
\newblock Unlearnable examples detection via iterative filtering.
\newblock In \emph{International Conference on Artificial Neural Networks}, pages 241--256. Springer, 2024{\natexlab{b}}.

\bibitem[Yu et~al.(2025{\natexlab{a}})Yu, Xia, Lin, Kong, Yang, Lu, Tan, and Kot]{yu2025towards}
Yi Yu, Song Xia, Xun Lin, Chenqi Kong, Wenhan Yang, Shijian Lu, Yap-Peng Tan, and Alex~C Kot.
\newblock Towards model resistant to transferable adversarial examples via trigger activation.
\newblock \emph{IEEE Transactions on Information Forensics and Security}, 2025{\natexlab{a}}.

\bibitem[Yu et~al.(2025{\natexlab{b}})Yu, Xia, Yang, Kong, Yang, Lu, Tan, and Kot]{yu2025mtlue}
Yi Yu, Song Xia, Siyuan Yang, Chenqi Kong, Wenhan Yang, Shijian Lu, Yap-Peng Tan, and Alex Kot.
\newblock Mtl-ue: Learning to learn nothing for multi-task learning.
\newblock In \emph{International Conference on Machine Learning}. PMLR, 2025{\natexlab{b}}.

\bibitem[Yuan and Wu(2021)]{ntga}
Chia-Hung Yuan and Shan-Hung Wu.
\newblock Neural tangent generalization attacks.
\newblock In \emph{International Conference on Machine Learning}, pages 12230--12240. PMLR, 2021.

\bibitem[Zhang et~al.(2019)Zhang, Gonzalez-Garcia, Weijer, Danelljan, and Khan]{updatenet}
L. Zhang, A. Gonzalez-Garcia, J. Weijer, M. Danelljan, and F. Khan.
\newblock Learning the model update for siamese trackers.
\newblock In \emph{ICCV}, 2019.

\bibitem[Zhang et~al.(2022)Zhang, Tian, Huang, Ye, Dai, Xie, and Tian]{zhang2022hivit}
Xiaosong Zhang, Yunjie Tian, Wei Huang, Qixiang Ye, Qi Dai, Lingxi Xie, and Qi Tian.
\newblock Hivit: Hierarchical vision transformer meets masked image modeling.
\newblock \emph{arXiv preprint arXiv:2205.14949}, 2022.

\end{thebibliography}
}

\end{document}